\documentclass{article}

\usepackage{PRIMEarxiv}
\usepackage[utf8]{inputenc} 
\usepackage[T1]{fontenc}    
\usepackage{hyperref}       
\usepackage{url}            
\usepackage{booktabs}       
\usepackage{amsfonts}       
\usepackage{nicefrac}       
\usepackage{microtype}      
\usepackage{lipsum}
\usepackage{fancyhdr}       
\usepackage{graphicx}       
\usepackage{mathtools}

\usepackage{subcaption}
\usepackage{amsmath, amssymb}
\usepackage{dblfloatfix}
\usepackage{natbib}
\usepackage{listings} 
\usepackage{array} 
\usepackage{multirow}
\usepackage{arydshln}
\usepackage{algorithm}
\usepackage{algpseudocode}
\usepackage{placeins}
\usepackage{doi}
\usepackage[dvipsnames]{xcolor}

\def\m#1{\texttt{#1}}
\newcommand{\tfc}{\emph{Traffic4cast }}

\newcommand{\IID}{\emph{i.i.d. }}
\newcommand{\covid}{COVID-19}

\hypersetup{
  colorlinks   = true, 
  urlcolor     = blue, 
  linkcolor    = blue, 
  citecolor   = blue 
}

\begin{document}

\title{Uncertainty Quantification for Image-based Traffic Prediction across Cities}


  \author{
  Alexander Timans\thanks{Corresponding author: \texttt{a.r.timans@uva.nl}; majority of work done while at ETH Zurich} \\
  Amsterdam Machine Learning Lab \\
  University of Amsterdam \\
  \And
  Nina Wiedemann \\
  Institute of Cartography and Geoinformation \\
  ETH Zurich \\
  \And
  Nishant Kumar \\
  Future Resilient Systems Programme \\
  Singapore-ETH Centre \\
  \And
  Ye Hong \\
  Institute of Cartography and Geoinformation \\
  ETH Zurich \\
  \And
  Martin Raubal \\
  Institute of Cartography and Geoinformation \\
  ETH Zurich \\
  }
  


 
   


\maketitle

\begin{abstract}

Despite the strong predictive performance of deep learning models for traffic prediction, their widespread deployment in real-world intelligent transportation systems has been restrained by a lack of interpretability. Uncertainty quantification (UQ) methods provide an approach to induce probabilistic reasoning, improve decision-making and enhance model deployment potential. 
To gain a comprehensive picture of the usefulness of existing UQ methods for traffic prediction and the relation between obtained uncertainties and city-wide traffic dynamics, we investigate their application to a large-scale image-based traffic dataset spanning multiple cities and time periods. We compare two epistemic and two aleatoric UQ methods on both temporal and spatio-temporal transfer tasks, and find that meaningful uncertainty estimates can be recovered. 
We further demonstrate how uncertainty estimates can be employed for unsupervised outlier detection on changes in city traffic dynamics. We find that our approach can capture both temporal and spatial effects on traffic behaviour in a representative case study for the city of Moscow. Our work presents a further step towards boosting uncertainty awareness in traffic prediction tasks, and aims to highlight the value contribution of UQ methods to a better understanding of city traffic dynamics. The code for our experiments is publicly available at 
\url{https://github.com/alextimans/traffic4cast-uncertainty}.

\end{abstract}

\keywords{uncertainty quantification, traffic prediction, unsupervised outlier detection, deep learning}



\section{Introduction}
\label{sec:intro}

Traffic prediction as part of wider mobility studies
forms a key aspect in the development of intelligent transportation systems for the modern city \citep{yin2021}. Accurate traffic predictions are useful for a range of applications in transportation systems and urban planning, allowing for improved traffic control and mitigation of traffic-related issues \citep{vlahogianni2004short, van1997recent}. However, traffic patterns appear highly uncertain due to their dependency on a wide range of volatile factors, including human decision-making \citep{zheng2020impact}. The complex and highly non-linear dynamics of traffic flow generally impede a perfect prediction of short-term traffic speed variations~\citep{nair2001non, li2004nonlinear}. In recent years, advances in data-driven methods such as deep neural networks (DNNs) have been adopted for traffic prediction tasks and tailored to address the specific challenges of its spatio-temporal nature with relative success \citep{tedjopurnomo2020}. 

While modern DNNs exhibit strong predictive performance in many applications, they have been shown to routinely overestimate their predictive power and exhibit overconfident behaviour \citep{guo2017calibration}. Furthermore, their lack of interpretability hampers perceived reliability and trustworthiness from a practitioner's perspective, and thus their integration into real-world application systems \citep{huang2020survey}. A central aspect of the problem is that a standard DNN will only provide predictive point estimates without incorporating notions of predictive uncertainty. Quantifying the uncertainty of a given model prediction can enhance model interpretability and provide a better-informed decision-making process for subsequent downstream tasks \citep{zhang2022}. For instance, uncertainty awareness has been shown to increase model usability for safety-critical applications in the medical domain \citep{laves2020} or in control tasks \citep{lutjens2019safe, richter2017safe}, and lead to improved human-machine interaction \citep{ayhan2020expert, roy2019, begoli2019}. Uncertainty quantification (UQ) is a valuable tool that may also prove beneficial in the domain of traffic prediction, particularly given the high predictive uncertainty expected to prevail for its challenging prediction tasks. Therefore, It is highly relevant to assess the effectiveness of UQ methods developed in the machine learning community for traffic prediction tasks.

Existing work on UQ for traffic prediction, such as \cite{mallick2022deep} or \cite{wu2021} (see \autoref{subsec:literature-traffic-pred}), relies heavily on loop counter data. While loop counters give accurate traffic observations with a high temporal resolution, they are constrained to a fixed and comparatively sparse set of locations primarily situated along major roads and within single urban regions \citep{jagadish2014big, pems}. This prevents a comprehensive picture of urban traffic dynamics, and it remains unclear how tested UQ approaches transfer across both space and time. Furthermore, there is a lack of analysis regarding the relations between obtained uncertainty estimates and city-wide traffic dynamics, as well as their practical relevance for possible downstream tasks. Our work addresses this gap by thoroughly comparing and analysing UQ methods on a large-scale image-based traffic dataset. We motivate that the selected data representation derived from dynamically moving probe vehicles allows us to comprehensibly analyse predictive uncertainty, and enables the application of UQ methods developed in the vision domain. We theoretically motivate and empirically assess four UQ methods suitable for image-based traffic prediction on both temporal and spatio-temporal transfer tasks. We subsequently leverage obtained uncertainty estimates for unsupervised outlier detection to demonstrate the value of quantifying uncertainty for downstream applications. 

Obtained results demonstrate that a proper combination of UQ methods can effectively capture predictive uncertainty that meaningfully relates to underlying traffic behaviour. We explicitly employ only widely applicable, distribution-free and readily implementable \emph{post-hoc}\footnote{i.e., they do not require modifications to the model training procedure} UQ methods, allowing for their straightforward adaptation by practitioners to different tasks and base models. We assess uncertainty estimation quality using similarly distribution-free approaches, such as conformal prediction intervals \citep{shafer2008}. 
Our work aims to showcase the usefulness of uncertainty-aware models for traffic prediction tasks and motivate the adoption of uncertainty quantification as a valuable tool for better-informed traffic analysis. To summarize, our contributions include the following key points:
\begin{itemize}
    \item a comparison of distribution-free, \emph{post-hoc} uncertainty quantification methods for a large-scale traffic prediction task, assessing their capabilities to produce meaningful uncertainty estimates in transfer tasks both across space, i.e., different cities, and time, i.e., different years;
    \item an analysis of obtained uncertainty estimates across spatial and temporal dimensions, linking them to the underlying city road network and traffic dynamics;
    \item demonstrating an application of obtained uncertainty estimates for unsupervised outlier detection, and that recovered outlier labels provide a meaningful signal both for changes in traffic dynamics across time and anomalous traffic situations as linked to city land use.
\end{itemize}

\section{Background and related work}
\label{sec:literature}

Sources of uncertainty in data-driven traffic prediction can be attributed to its two core components - the model and the data. Models such as DNNs might be sufficiently expressive to capture traffic complexity, but will suffer from a lack of available high-quality traffic measurements. Traffic data is characterized by the interaction of a large number of decision-making agents constrained by infrastructure properties, and the human component induces inherent uncertainty. These two facets align well with the distinction of two types of uncertainties found in the UQ literature. In the following subsections we introduce this distinction and follow with selected UQ methods for both types. We then discuss UQ for short-term traffic prediction, and establish a connection to unsupervised outlier detection, a key application we study.

\subsection{Epistemic and aleatoric uncertainty}
\label{subsec:literature-epi-alea}

The literature on uncertainty quantification commonly distinguishes between epistemic and aleatoric sources of uncertainty in a given problem setting \citep{kiureghian_aleatory_2009, kendall2017, depeweg2018}. Epistemic uncertainty - which is also referred to as model uncertainty - arises from potential misspecifications in the modelling process for the unknown data-generating process (DGP) the learning model aims to capture. In other words, it can be interpreted as the uncertainty over the approximation quality of the empirical risk minimizer $\hat{f}$ for the true risk minimizer $f^*$ based on the limited set of data samples at hand. 
In principle, this uncertainty is reducible under additional information on the DGP, as $\hat{f} \to f^{*}$ for increasing sample size $N \to \infty$ and a sufficiently rich hypothesis class. Aleatoric uncertainty - also known as data uncertainty - represents the uncertainty from sources of randomness that cannot be resolved even in light of additional information. This includes the inherent stochasticity in sampling from the DGP, as the relationship between inputs and targets for a given learning task is typically considered to be non-deterministic, i.e., following an unknown probability distribution. It also includes sources of data noise that cannot be practically mitigated, such as sensor measurement noise in the sampling procedure. It is therefore considered irreducible and can be properly quantified at best. 
In this work, we aim to quantify both epistemic and aleatoric uncertainties. See \cite{hullermeier2022} for a more in-depth discussion on their distinction.

\subsection{General uncertainty quantification methods}
\label{subsec:literature-uq-methods}

\begin{figure}
    \centering
    \includegraphics[width=0.95\columnwidth]{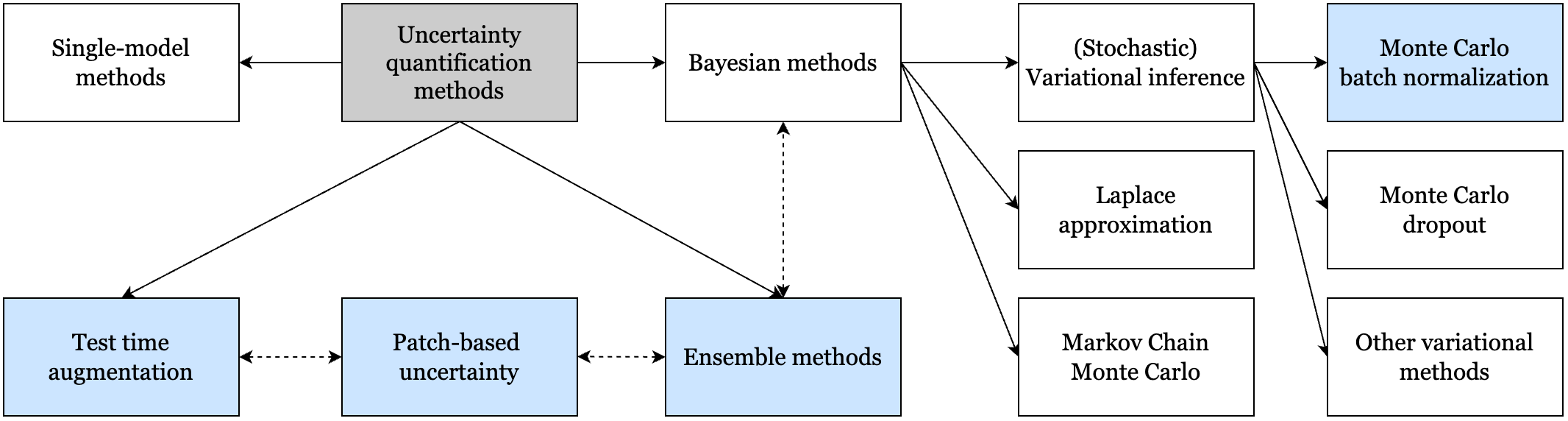}
    \caption{Distinction of research branches in the uncertainty quantification literature. Solid arrows denote the taxonomy, dashed arrows depict associations between methods which are detailed in \autoref{sec:uq-methods}. In particular, patch-based uncertainty is not part of the general literature. Methods implemented and tested in this study are highlighted in blue.}
    \label{fig:uqmethods}
\end{figure}

Recent traction in research on uncertainty has led to a multitude of UQ methods for DNNs that can be separated along multiple axes \citep{abdar2021}. We follow a recent survey by \cite{gawlikowski_survey_2022} and distinguish general research branches as shown in \autoref{fig:uqmethods}, placing our implemented UQ methods into context. Approaches for modelling epistemic uncertainty include (approximate) Bayesian methods, which assume distributions over a model's learnable parameters or some subset thereof \citep{blundell2015, ritter2018, kristiadi2020, daxberger2021}. These commonly require a modified training procedure (i.e., they are not \emph{post-hoc}) or impose some distributional assumptions. A prominent approach tailored to DNNs with dropout layers is Monte Carlo dropout \citep{gal2016dropout}, which has sparked further dropout variations \citep{mcclure2017, gal2017, mobiny2021}. Approaches following a similar idea for batch normalization layers are \cite{teye2018bayesian} and \cite{atanov2019uncertainty}. Using model ensembling to capture epistemic uncertainty has been championed by \citet{lakshminarayanan2017} with the introduction of deep ensembles. This has sparked a series of further model averaging proposals such as batch ensembles \citep{wen2020}, snapshot ensembles \citep{huang2017} or hyper ensembles \citep{wenzel2020}. 

Test time augmentation as an approach for quantifying aleatoric uncertainty in a \emph{post-hoc} manner has been introduced by \citet{ayhan2018test}, and has predominantly been used in image-to-image regression tasks such as medical imaging \citep{moshkov2020, wang2018automatic}. In the context of segmentation tasks, a particular form of data augmentation may be implemented, whereby the original image is partitioned into smaller patches for use in training, as opposed to using the image in its entirety~\citep{sreekanth2018novel, misra2020patch, schnurer2021detection}. This approach may be viewed as a method of uncertainty when the model is fed overlapping patches at test time, and the variance in predictions is recorded, which we refer to as patch-based uncertainty. While it is our understanding that this method has not been formally introduced, it has been employed in a limited number of studies in the field of medical imaging~\citep{klages2020patch, ghimire2020patch, zhang2006image} or soil pattern analysis~\citep{jenerette2006points}. 

Finally, single-model methods refer to models that are not ensembles and oftentimes yield uncertainty estimates in a single forward pass, increasing their efficiency at inference time. One such approach to modelling aleatoric uncertainty is achieved through the addition of learnable variance parameters and imposing a distributional form on model outputs, using e.g. Gaussian likelihoods \citep{kendall2017, chua2018deep} or assumed density filtering \citep{gast2018, loquercio2020general}. Other single-model methods based on non-parametric modelling include \cite{tagasovska2019, oala2020} and \cite{postels2019}. Most UQ methods aim to capture one type of uncertainty, and proposed frameworks to capture both types commonly combine two separate methods. 
For instance, \citet{kendall2017} combine Monte Carlo dropout and learnable variance parameters, \cite{chua2018deep} combine deep ensembling with learnable parameters, and \cite{wang2019aleatoric} combine Monte Carlo dropout with test time augmentation. Similarly, our framework in this work combines two approaches, and we motivate our selection of UQ methods in \autoref{sec:uq-methods}.

\subsection{Uncertainty quantification for traffic prediction}
\label{subsec:literature-traffic-pred}

The field of short-term traffic prediction has been an integral part of transportation research for several decades \citep{vlahogianni2014}. Like other research areas, it has benefitted from recent advances in modern DNNs, with models trained on large-scale data from loop counters and other sensors addressing some of the specific challenges of spatio-temporal traffic data (see \citet{yin2021, tedjopurnomo2020, ye2021} for recent surveys). Some recent applications include detecting traffic congestion and peak traffic conditions \citep{kumar2021, yu2017deep}, computing expected travel times \citep{wang2018will, pang2018learning} or quantifying passenger demand \citep{bai2019stg2seq, li2019learning, lee2018forecasting}. 

Only a limited number of traffic prediction approaches account for uncertainty. \citet{dejong2007} provide a partial review of parametric statistical models for traffic prediction, wherein predictive uncertainty is captured via prediction intervals. \citet{matas2012traffic} also quantify uncertainty via prediction intervals for a parametric regression model used on highway traffic data in Spain. \citet{GUO201450} extend statistical time series models with adaptive Kalman filters to predict short-term traffic flows, and similarly quantify uncertainty via prediction intervals. More recently, \citet{rodrigues2018heteroscedastic} use Gaussian processes to model predictive distributions for time series data on traffic speeds from road segments in Copenhagen
. These works however rely on the use of parametric models and require stronger modelling assumptions. Works employing DNNs include \citet{wu2021} who evaluate six different UQ techniques with the DCRNN model on the loop-counter dataset METR-LA \citep{jagadish2014big}. They find a Markov Chain Monte Carlo approach to provide the best results in comparison to other UQ methods, such as quantile regression or Monte Carlo dropout. In similar work, \citet{mallick2022deep} equip the DCRNN model with quantile regression and optimized hyperparameters to achieve improved uncertainty scores on the same dataset. \citet{wu2021bayesian} extend a graph network model to the Bayesian setting and quantify uncertainty for traffic speeds on the PeMS loop-counter dataset \citep{pems}. In other related work, \citet{lana2022measuring} combine different UQ methods - including deep ensembles, Monte Carlo dropout and conformal prediction - with different learning models on traffic flow readings from loop counters across the city of Madrid. They find conformal prediction with a random forest regressor to work best. Finally, an identified application of UQ to the \tfc dataset is \citet{maas2020}, who apply quantile regression with a Graph-WaveNet model to both METR-LA and a previous version of the \tfc dataset. However, they report a notable decrease in predictive performance compared to their baseline graph network model without UQ, and provide only a brief and qualitative evaluation of obtained uncertainties. We fill the open gap by providing a comprehensive analysis of practical, distribution-free UQ methods on this large-scale image-based traffic dataset, noting that related work relies almost exclusively on loop counter data.

\subsection{Unsupervised outlier detection}
\label{subsec:literature-outliers}

Outlier detection continues to be a prominent topic in traffic prediction, prompting recent surveys such as \citet{djenouri2019survey} for traffic flows or \citet{santhosh2021} with a focus on vision-based traffic data. Applications of different outlier detection methods - including unsupervised - to traffic prediction tasks entail \citet{chen2010comparison} for outliers in traffic flows and travel time prediction, \citet{li2009} for outliers in traffic speeds, \citet{pang2011mining} for outliers in driving patterns from GPS taxicab data, or \citet{shi2021} for stream-based trajectory traffic data. Unsupervised outlier detection methods are a class of methods that do not require outlier labels, but instead rely on varying assumptions on the data, such as on the probability of outliers (see \autoref{subsec:outliers-exp-setup}). A number of well-established approaches are based on statistical principles such as neighbourhood- and distance measures \citep{eskin2002geometric, breunig2000lof} or hypothesis testing \citep{beckman1983outlier, barnett1994outliers}, both for parametric and non-parametric density estimates \citep{goldstein2012histogram, bishop1994novelty, desforges1998applications}. In \autoref{sec:outlier-detection} we design such a hypothesis testing procedure using epistemic uncertainty estimates. Our outlier approach requires minimal data assumptions to maintain a distribution-free nature. Such a combination of distribution-free uncertainty estimates and outlier detection method for image-based traffic data is, to the best of our knowledge, novel in the recent traffic prediction literature.

\section{Uncertainty quantification methods}
\label{sec:uq-methods}

\begin{figure}
    \centering
    \includegraphics[width=0.95\columnwidth]{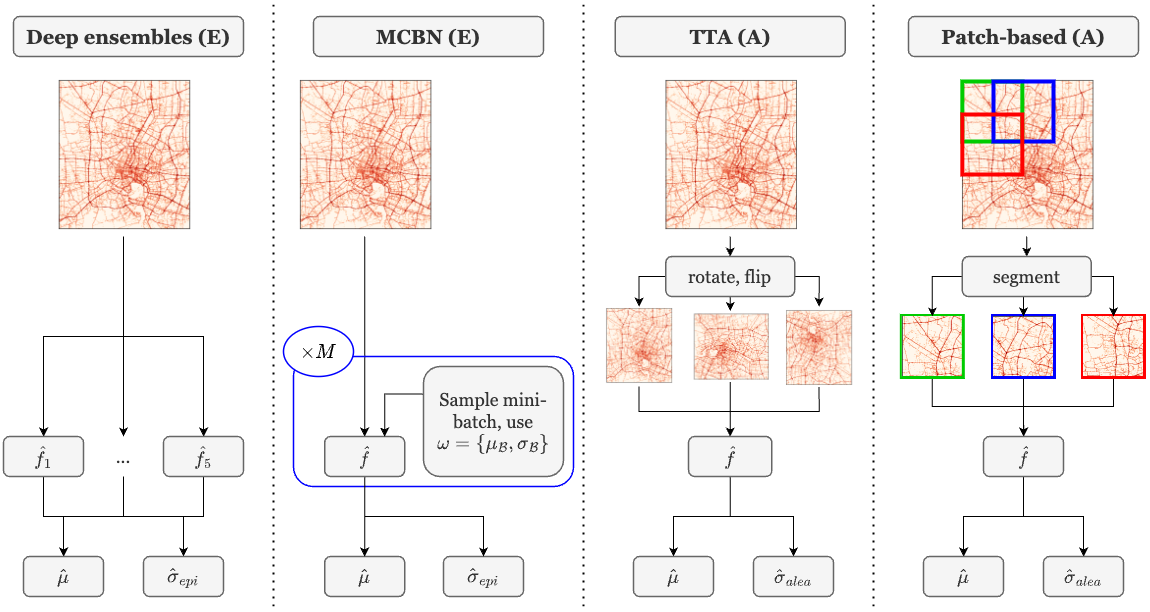}
    \caption{Schematic diagrams for the four implemented UQ methods on a given test image. Modelled uncertainty is denoted in brackets (\m{E}: epistemic, \m{A}: aleatoric). Each method provides a predicted point estimate and a corresponding uncertainty value after passing through the model $\hat{f}$. \m{MCBN} and \m{TTA} refer to Monte Carlo batch normalization and test time augmentation respectively.}
    \label{fig:uq_diagrams}
\end{figure}

In this study, we examine four methods of UQ: deep ensembles and Monte Carlo batch normalization (MCBN) for epistemic uncertainty, as well as test time augmentation (TTA) and patch-based uncertainty for aleatoric uncertainty. Our selection of methods is guided by the following primary criteria: 1) their lack of distributional assumptions on the data, which promotes their application potential for a wide range of tasks, 2) their suitability for image-to-image regression, and 3) their mostly \emph{post-hoc} quantification procedure, which allows them to be easily paired with various predictive base models. These properties are intended to facilitate their adaptation by practitioners to different prediction tasks and settings. Given these criteria we therefore do not consider some popular UQ methods such as those based on Bayesian learning. For each type of uncertainty, we compare a standard approach from the literature, namely deep ensembles and TTA, to lesser-known and more novel approaches, namely MCBN and patch-based uncertainty. We provide a brief theoretical motivation and description for each UQ method in the following. Since the selected methods for aleatoric UQ do not have a proper theoretical foundation in the literature, we outline our own theoretical justification for their use as uncertainty quantifiers. The four methods are visualized in \autoref{fig:uq_diagrams}. Additional implementation details can be found in \autoref{app:uq_implementation}.

\subsection{Deep ensembles}
\label{subsec:uq-methods-ensemble}

While \citet{lakshminarayanan2017} position deep ensembles as an alternative to Bayesian approaches, we follow recent work that motivates deep ensembles theoretically as an approximate Bayesian inference approach, or instance of Bayesian model averaging \citep{gustafsson2020a, lindqvist2020, wilson2020}. Given some training data $\mathcal{D} = \{ (x_i, y_i) \}_{i=1}^{N}$, the Bayesian framework aims to recover the predictive posterior distribution $p(y^* \mid x^*, \mathcal{D})=\int p(y^* \mid x^*, \theta)\,p(\theta \mid \mathcal{D})\,d\theta$ for a given test sample $(x^*, y^*)$. This requires determining a typically intractable posterior distribution $p(\theta \mid \mathcal{D}) \propto p(Y \mid X, \theta) \,\, p(\theta)$ over the model weights $\theta$ via Bayes' theorem. An approximation for $p(y^* \mid x^*, \mathcal{D})$ can be obtained using Monte Carlo samples $\{\theta_m\}_{m=1}^M$, which are drawn from an approximate posterior $q(\theta) \approx p(\theta \mid \mathcal{D})$. We thus approximate $p(y^* \mid x^*, \mathcal{D}) \approx \frac{1}{M} \sum_{m=1}^{M}\,p(y^* \mid x^*, \theta_m)\,,\,\, \theta_m \sim q(\theta)$, where the parameter setting $\theta_m$ represents a sampled point mass from $q(\theta)$. An ensemble with $M$ independently trained members results in $M$ such different parameter settings $\theta_m$, which can then be interpreted as \IID Monte Carlo samples from the approximate posterior $q(\theta)$ \citep{gustafsson2020a}. The approximation quality for $p(y^* \mid x^*, \mathcal{D})$ is primarily determined by the sample count, i.e. the ensemble size $M$. Perhaps surprisingly, even small ensembles have been found to empirically perform well \citep{chua2018deep, lang2022}. Following \citet{lakshminarayanan2017}, we use the recommended default ensemble size $M=5$ and induce diversity among ensemble members via random weights initialization and data shuffling. We take the mean over member predictions as a point estimate $\hat{\mu}$, and their standard deviation as an epistemic uncertainty estimate $\hat{\sigma}_{epi}$.

\subsection{Monte Carlo batch normalization}
\label{subsec:uq-methods-mcbn}

Following its introduction by \citet{ioffe2015batch}, batch normalization (BN) has seen wide success in neural network training and arguably replaced dropout as the default regularization technique in many recent model architectures \citep{ioffe2017, bjorck2018understanding}. Given training feature data $X$ and a randomly sampled mini-batch $\mathcal{B}=\{x_b\}_{b=1}^B$ thereof, a BN layer will normalize each sample $x_b$ in the mini-batch at training time as $BN^{train}(x_b) = \big((x_b - \mu_{\mathcal{B}})/\sqrt{\sigma^{2}_{\mathcal{B}} + \epsilon}\big)\,\gamma\,+\,\beta$ before feeding it to the next network layer. Parameters $\gamma$ and $\beta$ are learnable scale and shift parameters 
, while $\epsilon$ is added for numerical stability; $\mu_{\mathcal{B}}$ and $\sigma^{2}_{\mathcal{B}}$ are the mean and variance of the mini-batch respectively. At test time, the BN layer usually replaces these batch-dependent normalization statistics with the ones computed over the whole feature data $X$ for a given test sample $(x^*, y^*)$, that is $BN^{test}(x^*) = ((x_b - \mu_{X})/\sqrt{\sigma^{2}_{X} + \epsilon})\,\gamma\,+\,\beta$. 
Since the computation of batch statistics at training time exhibits stochasticity through the mini-batch sampling procedure, \citet{teye2018bayesian} propose to exploit this effect at test time to obtain epistemic uncertainty estimates, in an approach similar in spirit to Monte Carlo dropout by \citet{gal2016dropout}. Namely, batch-dependent statistics are also maintained at test time, and interpreted as \IID Monte Carlo samples to approximate the predictive posterior distribution $p(y^* \mid x^*, \mathcal{D})$. The approach dubbed `Monte Carlo batch normalization' can therefore be theoretically motivated as an approximate Bayesian inference method, similarly to deep ensembles. In particular, if we consider the extended set of learnable parameters in a model with BN as $\theta' = \{\theta, \gamma, \beta\}$ and decouple it from the stochastic statistics $\omega = \{\mu_{\mathcal{B}}, \sigma_{\mathcal{B}}\}$, we obtain an approximation as $p(y^* \mid x^*, \mathcal{D}) \approx \frac{1}{M} \sum_{m=1}^{M}\,p(y^* \mid x^*, \theta', \omega_m)\,,\,\, \omega_m \sim q_{\theta'}(\omega)$, where $q_{\theta'}(\omega)$ is an approximate posterior restricted to sampling randomness only for $\omega$. More detailed justifications can be found in \citet{teye2018bayesian}. The Monte Carlo samples are obtained by running $M$ forward passes with independently sampled mini-batches and respectively computed batch statistics. We run $M=10$ such forward passes for each test sample, and take the mean over these passes as a point estimate $\hat{\mu}$, and their standard deviation as an epistemic uncertainty estimate $\hat{\sigma}_{epi}$. While this UQ method is not entirely \emph{post-hoc} in the sense that it requires a model with BN layers, we still consider it practical due to the widespread adoption and ease of integration of BN. Furthermore, it may be easier to adopt than dropout for e.g. convolution-based models, for which it is unclear what dropout scheme is optimal \citep{poernomo2018biased}.

\subsection{Test time augmentation}
\label{subsec:uq-methods-tta}

Test time data augmentation has been used both for performance improvement and quantifying aleatoric uncertainty, and is primarily motivated from an empirical perspective \citep{ayhan2018test, moshkov2020, jin2018deep, chlap2021review}. A more formal mathematical formulation is suggested by \citet{wang2019aleatoric}, in which reversible data augmentations are interpreted as transformations of a latent variable. We build upon their idea and similarly suggest a theoretical formulation of TTA in which augmentations can be interpreted via a transformation operator on the data. In contrast to \citet{wang2019aleatoric} we neither have to specify a latent space nor parameterize an acquisition model but rather work directly with the stochasticity of the operator, providing a more straightforward motivation. We frame our argumentation in the context of an image-based regression task, albeit further generalization may be possible.

For a given input image $x \in \mathcal{X}$, any augmentations performed on $x$ can be summarized in a general transformation operator $\tau: \mathcal{X} \to \mathcal{X};\, x \mapsto \tau(x, \lambda)$ with transformation parameter $\lambda$. $\tau$ can either be a single augmentation, or a series of chained augmentations applied to $x$, and similarly, $\lambda$ can be a single scalar or a tuple denoting the parameters of each augmentation. We restrict these augmentations to be deterministic and spatially reversible, similarly to e.g. \citet{ayhan2018test, wang2019aleatoric}. We employ horizontal translations $f_h$, vertical translations $f_v$ and rotations $r$, but do not consider for instance hue or saturation changes. This is due to the one-to-one mapping of traffic information to pixel values in the \tfc dataset (see \autoref{subsec:uq-experiments-dataset}). Following their randomized application to $x$, these augmentations can be interpreted as random variables following individually specified distributions, namely $f_h \sim Bern(0.5),\,\,f_v \sim Bern(0.5),\,\,r \sim \mathcal{U}(\{\frac{\pi}{2},\,\pi,\,\frac{3\pi}{2},\,2\pi\})$, summarized in the random variable $\lambda = (f_h,\,f_v,\,r),\,\,\lambda \sim p(\lambda)$. Given its dependency on $\lambda$, any randomness is propagated such that $\tau \sim p(\tau \mid \lambda) \equiv p_\lambda(\tau)$ follows a general distribution not further specified. 

Given a training dataset $\mathcal{D}$ and assuming a non-deterministic relationship $p(y \mid x)$ for model inputs and outputs, a new test sample $(x^*,y^*)$ gives rise to a predictive distribution $p(y^* \mid x^*,\,\mathcal{D})$ for which we aim to capture some measure of aleatoric uncertainty. If we consider the application of $\tau$ to inputs at test time, we can express $p(y^* \mid x^*,\,\mathcal{D})$ in terms of a marginalization over $\tau$ as 
\begin{equation}
    p(y^* \mid x^*,\,\mathcal{D}) = \int p(y^* \mid x^*,\,\mathcal{D},\,\tau)\,d\tau = \int \hat{f}_\theta(\tau(x^*,\,\lambda))\,\,p_\lambda(\tau)\,d\lambda,
\end{equation}
where $\hat{f}_\theta(\cdot)$ denotes a forward-pass through the model with fixed weights $\theta$ trained on $\mathcal{D}$. Similarly to the methods for epistemic uncertainty, the presented integral can be approximated using \IID Monte Carlo samples drawn from the stochastic component $\tau$. In particular, the first two moments of $p(y^* \mid x^*,\,\mathcal{D})$ can be empirically approximated as 
\begin{equation}
    \hat{\mu}:=\,\mathbb{E}_{p(y^* \mid \cdot)}[y^*] = \mathbb{E}_{p_{\lambda}(\tau)}[\hat{f}_{\theta}(\tau(x^*,\,\lambda))] \approx \frac{1}{M}\sum_{m=1}^{M}\hat{f}_{\theta}(\tau_m(x^*,\,\lambda)),\,\,\tau_m \sim p_\lambda(\tau),
\end{equation}
\begin{equation}
    \hat{\sigma}^2 := \, \text{Var}_{p(y^* \mid \cdot)}[y^*] = \text{Var}_{p_{\lambda}(\tau)}[\hat{f}_{\theta}(\tau(x^*,\,\lambda))] \approx \frac{1}{M}\sum_{m=1}^{M}\hat{f}_{\theta}(\tau_m(x^*,\,\lambda))^T \hat{f}_{\theta}(\tau_m(x^*,\,\lambda)) - \hat{\mu}^T\hat{\mu}, \,\, \tau_m \sim p_\lambda(\tau).
\end{equation}

We can then interpret $\hat{\mu}$ as a predicted point estimate, and $\sqrt{\hat{\sigma}^2} \equiv \hat{\sigma}_{alea}$ as a measure of aleatoric uncertainty, since the stochastic transformation operator $\tau$ is applied to the input domain and thus directly relates to data uncertainty \citep{kendall2017}. In practice, samples from $p_\lambda(\tau)$ are constructed by sampling from each augmentation and returning a sampled parameter tuple $\lambda$. For example, an unaugmented image can be interpreted as augmented with no flips and 360° rotation, i.e. $\tau(\cdot\,,\lambda)$ with $\lambda = (0, 0, 2\pi)$.

\subsection{Patch-based uncertainty}
\label{subsec:uq-methods-patches}

\begin{figure}
    \centering
    \includegraphics[width=0.45\columnwidth]{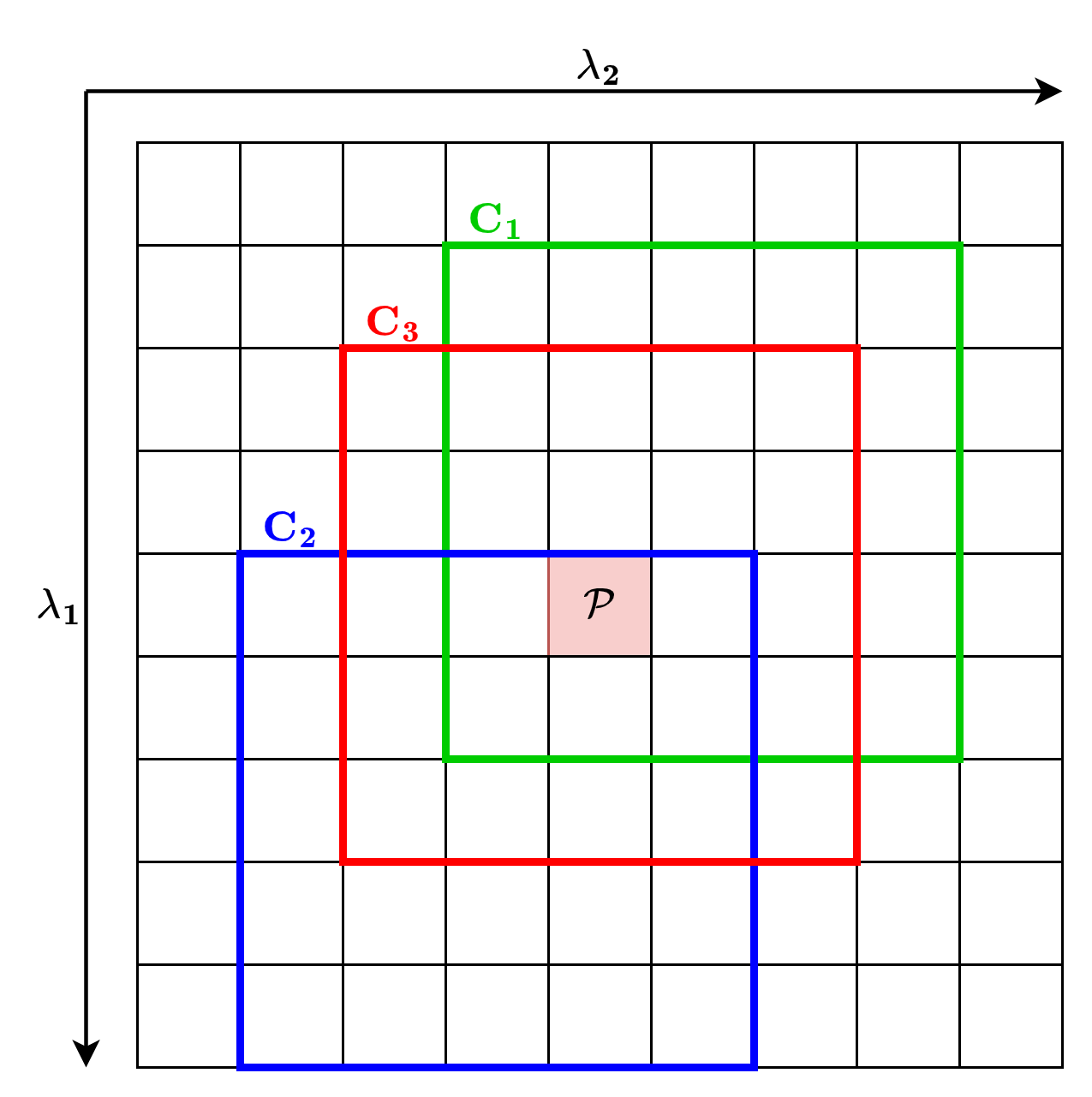}
    \caption{Exemplary gridded image of size $9 \times 9$ with three sampled context patches $C_1,C_2,C_3$ each of size $5 \times 5$, corresponding to different translations $\lambda_1, \lambda_2$ around $\mathcal{P}$.}
    \label{fig:theory-patch-ex}
\end{figure}

Patch-based processing and its use for uncertainty quantification has seen little theoretical description in previous referenced work. In the setting of an image-to-image regression task, we propose a theoretical motivation for patch-based uncertainty wherein we interpret an image pixel's spatial \emph{context} as a random variable from which \emph{context patches} can be sampled. We regard the spatial context as an additional model input that relates its stochasticity to the data, thus providing an aleatoric uncertainty measure \citep{kendall2017}. 

We consider a single image pixel $\mathcal{P}$ at a fixed location $(i, j)$, which lies within the bounds of a gridded input image with height $I_h$ and width $I_w$. We denote the spatial context around $\mathcal{P}$ as $\mathcal{C}$, and restrict $\mathcal{C}$ to quadratic patches of size $d \times d, \,\, d \in \{0,\dots,\min\{I_h,I_w\}\}$ to simplify processing. $\mathcal{C}$ defines a neighbourhood of pixels included in the patch and is parametrized by vertical and horizontal translations $\lambda = (\lambda_1, \lambda_2)$ centered around $\mathcal{P}$. For $d$ uneven\footnote{the same case can be made for $d$ even by shifting the values of $C$ accordingly} we can define any realized context patch $C$ of the spatial context $\mathcal{C}$ as having the following shape: 
\begin{equation}
    C = \{i+\lambda_1-\left\lfloor \frac{d}{2} \right\rfloor,\, \dots, \,i+\lambda_1+\left\lfloor \frac{d}{2} \right\rfloor\} \times \{j+\lambda_2-\left\lfloor \frac{d}{2} \right\rfloor,\, \dots, \,j+\lambda_2+\left\lfloor \frac{d}{2} \right\rfloor\},
\end{equation}
\begin{equation}
    C \subset \{k \in \mathbb{N} \mid 0 < k \le I_h\} \times \{k \in \mathbb{N} \mid 0 < k \le I_w\}.    
\end{equation}
The domain of vertical and horizontal translations is given by $\lambda_1, \lambda_2 \in \mathbb{Z}_\lambda := \{-\left\lfloor \frac{d}{2} \right\rfloor,\, \dots,\,0,\,\dots,\left\lfloor \frac{d}{2} \right\rfloor\}$. We provide an example in \autoref{fig:theory-patch-ex}, where we consider a gridded image of size $I_h = I_w = 9$ and set $d=5,\,(i,j) = (5,5)$. For $\lambda = (-1, 1)$ we obtain context patch $C_1 = \{2,\,\dots,\,6\} \times \{4,\,\dots,\,8\}$, while for $\lambda = (2,-1)$ we get $C_2 = \{5,\,\dots,\,9\} \times \{2,\,\dots,\,6\}$. Setting $\lambda=(0,0)$ recovers the context patch $C_3 = \{3,\,\dots,\,7\} \times \{3,\,\dots,\,7\}$ centered at $\mathcal{P}$. We consider the sampling of random translations $\lambda_1, \lambda_2$ uniformly from their realizable values, i.e. $\lambda_1, \lambda_2 \sim \mathcal{U}(\mathbb{Z}_\lambda)$, and extend the random variable interpretation to the spatial context $\mathcal{C}$ via a general distribution, i.e. $p(\mathcal{C} \mid \lambda) = p(\mathcal{C} \mid \lambda_1, \lambda_2) \equiv p_{\lambda}(\mathcal{C})$. Let $\hat{f}_{\theta}$ be a trained prediction model that receives the pixel value $x^*$ and the spatial context $\mathcal{C}$ for $\mathcal{P}$ as inputs at test time. A predictive distribution on the pixel output $y^*$ is then given by 
\begin{equation}
    p(y^* \mid x^*,\,\mathcal{D}) = \int p(y^*\mid x^*,\,\mathcal{D},\,\mathcal{C})\,\,d\,\mathcal{C} = \int \hat{f}_{\theta}(x^*, \mathcal{C}) \,p_{\lambda}(\mathcal{C}) \,d\lambda, 
\end{equation} 
where $\hat{f}_{\theta}(\cdot)$ denotes a forward-pass through the model with fixed weights $\theta$. Similarly to the other presented UQ methods, the above integral is approximated using Monte Carlo samples. Given $M$ \IID draws for $\lambda_1, \lambda_2$ resulting in $M$ randomly sampled context patches, we feed the pixel input $x^*$ and context patches $C_1,\dots,C_M$ through the model and approximate the first two moments of $p(y^* \mid x^*,\,\mathcal{D})$ as follows: 
\begin{equation}
    \hat{\mu}:=\,\mathbb{E}_{p(y^* \mid \cdot)}[y^*] = \mathbb{E}_{p_{\lambda}(\mathcal{C})}[\hat{f}_{\theta}(x^*, \mathcal{C})] \approx \frac{1}{M}\sum_{m=1}^{M}\hat{f}_{\theta}(x^*, C_m), 
\end{equation}
\begin{equation}
    \hat{\sigma}^2 := \, \text{Var}_{p(y^* \mid \cdot)}[y^*] = \text{Var}_{p_{\lambda}(\mathcal{C})}[\hat{f}_{\theta}(x^*, \mathcal{C})] \approx \frac{1}{M}\sum_{m=1}^{M}\hat{f}_{\theta}(x^*, \mathcal{C})^T \hat{f}_{\theta}(x^*, \mathcal{C}) - \hat{\mu}^T\hat{\mu}. 
\end{equation}
We then take $\hat{\mu}$ as the predicted point estimate, and $\sqrt{\hat{\sigma}^2} \equiv \sigma_{alea}$ as an aleatoric uncertainty estimate for pixel $\mathcal{P}$. Repeating this procedure yields pixel-level estimates for all pixels in a given test image, providing a general estimation of heteroscedastic (i.e. variable) aleatoric uncertainty. In practice, we follow the approach by \citet{wiedemann2021traffic} 
and use a sliding window to efficiently extract patches at test time. For more details we refer to \autoref{app:patches}.

\subsection{Predictive uncertainty}
\label{subsec:uq-methods-predictive}

The characterisation of uncertainty arising from different sources of randomness as either epistemic or aleatoric depends on the specific problem setting, as initially observed by \citet{kiureghian_aleatory_2009}. 
Assuming a fixed and well-defined problem setting (i.e., sources of randomness are identified and do not change), a model's total predictive uncertainty for a given point estimate can be additively separated into its aleatoric and epistemic components via the law of total variance \citep{deisenroth2020mathematics, depeweg2018}. Given a test sample $(x^{*},y^{*})$, if we consider a prediction model $\hat{f}_{\theta}$ trained on data $\mathcal{D}$ outputting both a mean estimate $\mathbb{E}[y^* \mid x^*, \theta]$ and variance estimate $\text{Var}(y^* \mid x^*, \theta)$, it holds for the predictive uncertainty that 
\begin{equation}
    \text{Var}(y^* \mid x^*, \mathcal{D}) = \mathbb{E}_m[\text{Var}(y^* \mid x^*, \theta)] \,+\, \text{Var}_m(\mathbb{E}[y^* \mid x^*, \theta]).
\end{equation}
Aleatoric uncertainty is quantified by the first component, and epistemic by the second, while $m \in \{1,\dots,M\}$  identifies the index of samples over the \emph{model domain}, obtained for example via multiple forward passes. In our case, the application of \emph{post-hoc} UQ methods means that $\hat{f}_{\theta}$ only outputs a mean estimate $\mathbb{E}[y^* \mid x^*, \theta]$. Therefore, the aleatoric uncertainty component is also expressed as a variation estimate $\text{Var}_i(\mathbb{E}[y^* \mid x^*, \theta])$, where $i$ now crucially identifies samples over the \emph{input domain}. 
We presume that the additive decomposition of total predictive uncertainty holds equivalently. 


\section{Uncertainty quantification experiments on the \tfc dataset}
\label{sec:uq-experiments}



Our experimental results in this section are organized as follows. Firstly, we describe the \tfc dataset and our experimental design, and formalize the metrics we use to assess uncertainty estimation quality. Proposed UQ methods are then compared quantitatively in \autoref{subsec:uq-experiments-comparison}, and obtained uncertainties are analysed spatially (\autoref{subsec:uq-experiments-spatial}) and temporally (\autoref{subsec:uq-experiments-temporal}).
Additional results can be found in \autoref{app:results_table}.

\subsection{\tfc dataset}
\label{subsec:uq-experiments-dataset}

Our analysis and experiments are based on the dataset provided by Here Technologies\footnote{\url{https://www.iarai.ac.at/traffic4cast/2021-competition/challenge/\#data} ; \url{https://www.here.com}} for the \tfc 2021 challenge~\citep{eichenberger2022traffic4cast}. The data is derived from trajectories of raw GPS position fixes, consisting of latitude, longitude, timestamp, recorded vehicle speed and driving direction. These were collected by a large fleet of probe vehicles driving on the roads of ten major international cities, from which we consider eight in this study: Antwerp, Bangkok, Barcelona, Moscow, Berlin, Chicago, Istanbul and Melbourne. For the first four cities, data is made fully available across two six-month periods in 2019 and 2020, i.e., 02/01/2019 to 30/06/2019 and 02/01/2020 to 30/06/2020. For the latter four cities, data is made available for the 2019 period only. 

The raw GPS data has been transformed into a visual representation which permits the use of image-based learning models. Each city is compressed into a gridded image representation of $495 \times 436$ pixels, where each image pixel corresponds to a spatial region of around $100\text{m} \times 100\text{m}$. For each pixel, the recorded volume and speed values are aggregated and normalized to a value in $[0,255]$\footnote{\url{https://www.iarai.ac.at/traffic4cast/forums/topic/competition-data-details/}; access needs registration} for each of four separate heading directions (NE, SE, SW, NW), resulting in eight image channels that extend a city's image representation to size $495 \times 436 \times 8$. The temporal sampling frequency of the probe vehicles is 5 min, i.e., 12 measurements per hour, leading to $12 \cdot 24 = 288$ temporally coherent samples per one full day of data. For each given six-month period of data, we thus obtain a total of $180 \cdot 288 = 51840$ temporal samples per city. Each temporal sample is the equivalent of such a city-wide image representation of traffic volume and speeds (see e.g. \autoref{fig:uq_diagrams} for a sample image for Bangkok). 

The \tfc dataset provides two key benefits over traditional loop-counter datasets for recording traffic behaviour. Firstly, it spans both multiple cities and time periods, allowing for large-scale and comprehensive evaluation of methods across both space and time. Secondly, sampling is not location-restricted and highly dynamic, allowing us to generate a dense city-wide traffic representation which is especially useful for analysing spatial traffic patterns. 
In addition, the image-based data representation has been successfully used in several iterations of the \tfc challenge, highlighting its ability to capture complex spatio-temporal dynamics \citep{kreil2020surprising, eichenberger2022traffic4cast}.

\subsection{Experimental setup}
\label{subsec:uq-experiments-setup}

The prediction task given by the \tfc 2021 challenge is as follows: given one hour of traffic data input, i.e., 12 temporally coherent samples, predict the traffic volumes and speeds for six future time steps, namely 5, 10, 15, 30, 45, and 60 min ahead. In other words, given a data input sample $x$ of size $12 \times 495 \times 436 \times 8$, provide a multi-output regression prediction $\hat{y}$ of size $6 \times 495 \times 436 \times 8$. In our experiments, we limit ourselves to predicting 60 min ahead, since the most distant time step is arguably the hardest to predict, but obtained results should translate well to shorter horizons.
Similarly, we primarily focus our evaluation on traffic speeds since they exhibit both higher magnitude and variability in their values than traffic volume.

Predicted point estimates are complemented by uncertainty estimates for each implemented UQ method. This is done per pixel and per channel, which we also refer to as a \emph{cell}. That is, for a given pixel $(i,j)$ and channel $c$, the tuple $(i,j,c)$ constitutes a cell. We thus fully retain the data’s channel dimension, i.e., directional traffic speed and volumes. Our prediction task then summarizes as follows: given an input sample $x$ of size $12 \times 495 \times 436 \times 8$, predict both values and uncertainties for the 60 min ahead time step. Our model output $\hat{y}$ for one sample is then of size $2 \times 495 \times 436 \times 8$. 

Available data is split into training, validation and test sets. 
In line with evaluating the generalisability of UQ methods across both space and time, the splits are motivated by considering temporal and spatio-temporal transfer tasks. Each model is trained only on the combined 2019 data across all cities, excluding Antwerp. The first half of 2020 data (available for Antwerp, Bangkok, Barcelona and Moscow) is used for validation and the latter half for testing. 
This way, the model has to adapt to any possible temporal domain shifts for all four cities. In particular, note that the 2019 data is pre-\covid~times, while the 2020 data is partially within \covid~period. For Antwerp, the model has to additionally adapt to a potential spatial domain shift, since it is reasonable to assume that traffic patterns may differ between cities in some form.

Regarding a choice of model architecture, we note that U-Net-based methods have been highly successful in the \tfc challenge, to the extent that a default U-Net architecture as proposed by \cite{ronneberger2015u} is provided to participants as a baseline method \footnote{\url{https://github.com/iarai/NeurIPS2021-traffic4cast/blob/master/baselines}}. The U-Net model is generally popular for image-to-image tasks, such as in the medical domain \citep{siddique2021}, and has spawned follow-up architectures, such as the U-Net++ \citep{zhou2019unetplusplus}. We therefore evaluate the performance of UQ methods on both the U-Net and U-Net++ baseline models, which we can confidently consider as having adequate baseline performance. Both architectures are slightly modified to accommodate batch normalization. Other popular DNN architectures for traffic prediction are not considered in this study due to several reasons: 1) they introduce additional modelling complexity and associated assumptions, such as the construction of road graphs for graph-based methods \citep{ye2021}; 2) they require involved model training, hyperparameter tuning and/or high computational budget, such as DCRNN \citep{li2017diffusion}; and 3) they lack the extensive evaluations within a standardized competition framework like the \tfc challenge, which can be considered an empirical `seal of approval' for our task. While the simplicity and adaptability of the U-Net's convolutional architecture contribute to its appeal, the UQ methods evaluated in this study are purely \emph{post-hoc} 
and can be utilized in conjunction with any reasonable model of preference. Model hyperparameter settings have been previously explored by \cite{eichenberger2022traffic4cast} for performance tuning. To maintain the broadest level of generality, we adopt their default values and do not perform additional tuning with regard to uncertainty estimation quality. Optimizing the model or parameter selection did not constitute part of the research focus of this work. The validation data is therefore used primarily as a calibration set for prediction interval construction, as detailed in the next section.

\subsection{Metrics}
\label{subsec:uq-experiments-metrics}

Calibration and sharpness are two important concepts related to the quality of uncertainty estimation \citep{gneiting2007a, naeini2015obtaining}. To capture both notions in a distribution-free manner, we utilize expected calibration scores and rank correlation for calibration, and prediction intervals via conformal prediction to capture sharpness. While there is no standardized set of metrics for assessing uncertainty estimation quality, other commonly used scores in regression tasks include predictive log-likelihood or CRPS \citep{nado2021uncertainty, chung2021uncertainty}. However, their practical implementation necessitates distributional assumptions, which is why we do not consider them here. We additionally measure predictive performance via mean squared error (MSE). Note that we are not primarily interested in achieving a lower MSE score. Instead, we use these scores to validate that the implemented UQ methods do not harm the model's overall predictive performance.

\subsubsection{Calibration}
\label{subsubsec:uq-experiments-metrics-calibration}

Calibration is the general notion that a model’s confidence in its predictions should match its predictive performance, e.g. by matching class assignment probabilities to its actual accuracy scores in a classification task \citep{guo2017calibration}. 
Similarly to e.g. \cite{levi2022, laves2020}, we extend the concept to regression tasks by considering a model well-calibrated when the model’s prediction error, as calculated via MSE, matches the predicted uncertainty, as calculated via a measure of variation. 
More formally, for some test data $\mathcal{D}^* = \{ (x^*_t, y^*_t) \}_{t=1}^{T}$ and a model with point and uncertainty predictions $\hat{\mu}(\cdot)$ and $\hat{\sigma}(\cdot)$ we want for any uncertainty value $\sigma$ that $\mathbb{E}_{\mathcal{D}^*}[(\hat{\mu}(x^*)-y^*)^2 \mid \hat{\sigma}(x^*)=\sigma]=\sigma$. 
We evaluate calibration performance by computing the expected normalized calibration error (ENCE) from \cite{levi2022}, given by 
\begin{equation}
    \text{ENCE} = \frac{1}{T}\sum_{t=1}^{T}\frac{|\hat{\sigma}(x^*_t)-\sqrt{(\hat{\mu}(x^*_t)-y^*_t)^2}|}{\hat{\sigma}(x^*_t)},    
\end{equation}
where $T=|\mathcal{D}^*|$ denotes the size of the test set\footnote{Note that this refers to the number of test samples on a \emph{per-cell} basis. Scalars can then be computed by additionally averaging across channels and spatial dimensions (for results tables) or channels only (for spatial maps).}. A drawback of this metric is its high sensitivity to the scale of $\hat{\sigma}$ due to normalization. We therefore additionally report Spearman’s rank correlation coefficient $\rho_{sp}$ between prediction residuals and uncertainties \citep{zar2014spearman}. It is non-parametric, restricted to values in $[-1, 1]$, and intuitively captures the notion of calibration defined above: a high correlation indicates a strong positive relationship between error and uncertainty, suggesting a well-calibrated model. If we consider the sets $E=\{|\hat{\mu}(x^*_t)-y^*_t|\}_{t=1}^T$ and $U=\{\hat{\sigma}(x^*_t)\}_{t=1}^T$ and denote the ranks of respective values as $R(\cdot)$, then the rank correlation can be computed as 
\begin{equation}
    \rho_{sp}=\frac{\text{Cov}(R(E),\,R(U))}{\sigma_{R(E)}\,\sigma_{R(U)}}.
\end{equation}

\subsubsection{Sharpness}
\label{subsubsec:uq-experiments-metrics-pi}

The notion of sharpness stipulates that predictions should preferably be made with high confidence whilst maintaining accuracy. In a regression setting, this naturally motivates the use of prediction intervals (PIs): given an interval constructed using uncertainty estimates, we prefer intervals that have high empirical coverage whilst maintaining a small width, i.e. they are `sharply' located around ground truth values. Typical metrics include mean prediction interval width and prediction interval coverage probability \citep{kompa2021, gawlikowski_survey_2022}. 

A promising framework to generate distribution-free prediction intervals in a \emph{post-hoc} manner is conformal prediction, which has recently promoted its applicability to DNNs \citep{vovk2005algorithmic, shafer2008, angelopoulos2023conformal}. Conformal prediction provides a method to construct PIs with theoretical marginal coverage guarantees under relaxed \IID assumptions. An empirical quantile $\hat{q}$ is selected from computed non-conformity scores which encode agreement between predictions and ground truths on a set of unseen `calibration' samples $\mathcal{D}^{cal}$ \footnote{the term `calibration' holds a different interpretation in the conformal literature}.
For a test sample $(x^*, y^*)$, a symmetrical prediction interval can then be constructed as $\mathcal{I}(x^*)=[I_l(x^*),\,I_u(x^*)]=[\hat{\mu}(x^*)-\hat{\sigma}(x^*)\,\hat{q},\,\,\hat{\mu}(x^*)+\hat{\sigma}(x^*)\,\hat{q}]$. Crucially, $\hat{q}$ is chosen such that we can guarantee marginal coverage for some user-defined error $\alpha$ across calibration and test sets, that is, we guarantee $\mathbb{P}_{\mathcal{D}^{cal}, \mathcal{D}^*}(y^* \in \mathcal{I}(x^*)) \ge 1-\alpha$. Intuitively, $\hat{q}$ can be considered an uncertainty scaling factor that corrects for the desired coverage level. Since coverage levels are fixed \emph{a priori}, we compare methods in terms of their mean PI width (MPIW) only: 
\begin{equation}
    \text{MPIW}=\frac{1}{T}\sum_{t=1}^{T}(I_u(x^*)-I_l(x^*))=\frac{1}{T}\sum_{t=1}^{T}\,2\,\hat{\sigma}(x_t^*)\,\hat{q}.    
\end{equation}
We prefer a lower MPIW for a given coverage level, which we fix to be 90\% (i.e. $\alpha = 0.1$). We validate the correct implementation of conformal prediction intervals in \autoref{app:coverage}.

\subsection{Comparison of uncertainty quantification methods}
\label{subsec:uq-experiments-comparison}

\begin{table*}[t]
\centering
\resizebox{\textwidth}{!}{

\begin{tabular}{l|rrrrr|rrrrr}

\toprule
{\textbf{City}} & \multicolumn{5}{c}{Antwerp (spatio-temporal transfer)} & \multicolumn{5}{c}{Moscow (temporal transfer)} \\

\midrule

\textbf{UQ method} & MSE $\downarrow$ & Uncertainty & MPIW $\downarrow$ & ENCE $\downarrow$ & $\rho_{sp}$ $\uparrow$
& MSE $\downarrow$ & Uncertainty & MPIW $\downarrow$ & ENCE $\downarrow$ & $\rho_{sp}$ $\uparrow$ \\

\midrule

\m{Ens (E)} & 80.28 & 0.258 $\pm$ 0.253 & 1.697 & 0.526 & 0.791 & \textbf{189.30} & 0.550  $\pm$ 0.296 & 16.040 & 2.367 & 0.641 \\
$\hookrightarrow$ \m{zero-mask}  & 296.52 & 0.733 $\pm$ 0.671 & 5.401 & 1.159 & 0.876 & \textbf{309.99} & 0.858 $\pm$ 0.449 & 26.098 & 3.898 & 0.673 \\

\m{MCBN (E)} & \textbf{77.35} & 0.147 $\pm$ 0.124 & 2.990 & 1.652 & 0.630 & 197.21 & 0.477 $\pm$ 0.264 & 21.820 & 6.606 & 0.593 \\
$\hookrightarrow$ \m{zero-mask}  & \textbf{285.69} & 0.350  $\pm$ 0.315 & 9.369 & 3.655 & 0.691 & 322.88 & 0.744 $\pm$ 0.408 & 35.236 & 10.20 & 0.603 \\

\midrule

\m{TTA (A)} &  79.24 & 0.191 $\pm$ 0.156 & 1.887 & 1.093 & 0.447 & 198.04 & 0.729 $\pm$ 0.312 & 16.960 & 2.950 & 0.450 \\
$\hookrightarrow$ \m{zero-mask}  & 292.81 & 0.512 $\pm$ 0.428 & 6.122 & 2.099 & 0.725 & 324.24 & 1.121 $\pm$  0.470 & 27.435 & 4.407 & 0.552 \\

\m{Patches (A)} & 77.43 & 0.066 $\pm$ 0.071 & 3.266 & 6.142 & 0.734 & 193.73 & 0.172 $\pm$ 0.108 & 19.510 & 19.160 & 0.637 \\
$\hookrightarrow$ \m{zero-mask}  & 286.12 & 0.182 $\pm$ 0.189 & 10.959 & 10.540 & 0.811 & 317.24 & 0.269 $\pm$ 0.166 & 31.693 & 29.850 & 0.627 \\

\midrule

\m{TTA+Ens (P)} & 80.71 & 0.831 $\pm$ 0.478 & 1.40 & 0.438 & 0.860 & 192.19 & 1.996 $\pm$ 0.669 & \textbf{14.050} & \textbf{0.846} & 0.658 \\
$\hookrightarrow$ \m{zero-mask}  & 298.12 & 1.787 $\pm$ 1.237 & 4.424 & 0.534 & 0.894 & 314.73 & 2.907 $\pm$ 0.982 & \textbf{22.859} & \textbf{1.054} & 0.714 \\

\m{Patches+Ens (P)} & 81.74 & 0.321 $\pm$ 0.284 & \textbf{1.148} & \textbf{0.239} & \textbf{0.938} & 211.34 & 1.251 $\pm$ 0.465 & 14.540 & 0.848 & \textbf{0.845} \\
$\hookrightarrow$ \m{zero-mask}  & 302.10 & 0.958 $\pm$ 0.798 & \textbf{3.688} & \textbf{0.520} & \textbf{0.941} & 346.09 & 1.985 $\pm$ 0.711 & 23.683 & 1.399 & \textbf{0.805} \\

\m{CUB (P)} & 79.01 & 0.565 $\pm$ 0.0 & 3.905 & 1.002 & -0.037 & 197.30 & 1.684 $\pm$ 0.0 & 14.760 & 1.325 & 0.002 \\
$\hookrightarrow$ \m{zero-mask}  & 291.94 & 1.800 $\pm$ 0.0 & 13.436 & 1.035 & -0.033 & 323.02 & 2.687 $\pm$ 0.0 & 23.846 & 1.479 & 0.007 \\

\bottomrule

\end{tabular}
}
\vspace{1mm}
    \caption{Comparison of UQ methods for Antwerp and Moscow in terms of traffic speed prediction. Predictions are obtained using a U-Net++ model. Methods are grouped by modelled uncertainty (\m{E}: epistemic, \m{A}: aleatoric, \m{P}: predictive). Scores are calculated both across all cells and only those cells with traffic activity, i.e., non-zero volume (\m{zero-mask}). Best scores per city, metric and masking across all methods are marked in \textbf{bold}. Arrows indicate favoured score direction (higher $\uparrow$, lower $\downarrow$).}
    \label{tab:uq_comparison}
\end{table*}

We evaluate the proposed uncertainty quantification methods on samples from the test split and compare them in \autoref{tab:uq_comparison}. As mentioned in \autoref{subsec:uq-experiments-setup}, we evaluate on Antwerp for spatio-temporal transfer capabilities. From the remaining three cities with test data, we report Moscow as a proxy for temporal transfer because it exhibits the highest traffic activity, as measured by total traffic volume. Since an image-based traffic representation includes spatial areas outside the city's road network, we also compute metrics by masking for cells that exhibit traffic activity across test samples, i.e., have a non-zero traffic volume (\m{zero-mask}). This allows for assessing the impact of data sparsity on obtained scores. 

We observe in \autoref{tab:uq_comparison} that while deep ensembles (\m{Ens}) expresses uncertainties of larger magnitude, these are both better calibrated and sharper on both tasks, favouring it over Monte Carlo batch normalization (\m{MCBN}) for epistemic uncertainty. For aleatoric uncertainty, test time augmentation (\m{TTA}) produces larger uncertainties than the patch-based approach (\m{Patches}) and scores better in terms of MPIW and ENCE, but not $\rho_{sp}$. Furthermore, as motivated in \autoref{subsec:uq-methods-predictive}, we combine two UQ methods modelling different types of uncertainty to obtain a measure of total predictive uncertainty, that is, we have $\hat{\sigma}_{pred} = \hat{\sigma}_{alea} + \hat{\sigma}_{epi}$. We consider using TTA with deep ensembles (\m{TTA+Ens}), and patch-based uncertainty with deep ensembles (\m{Patches+Ens}). 
Additionally, we evaluate a constant uncertainty baseline (\m{CUB}), wherein we assign each cell the standard deviation in model predictions across test samples as a fixed uncertainty value. It can be considered an informed estimate determined by the variation in test sample ground truths, as captured by the variation in test sample predictions. The idea is similar to e.g. \cite{teye2018bayesian} who optimize a constant value on validation data, and aims to provide context to obtained predictive uncertainty estimates. 

We see that \m{CUB} provides reasonable scores both for MPIW and ENCE. In particular, the magnitude of uncertainties is notably larger than for any UQ method modelling only one type of uncertainty, which is expected. However, both \m{TTA+Ens} and \m{Patches+Ens} are able to outperform \m{CUB} in terms of uncertainty quality, albeit perhaps at a smaller margin than anticipated. Crucially, combining UQ methods results in higher-quality uncertainty estimates than any single UQ method, suggesting that we are able to more fully and accurately capture existing uncertainty in our prediction task. In direct comparison, there is no clear favourite for modelling such predictive uncertainty, although \m{Patches+Ens} arguably scores slightly higher. However, we observed patch-based uncertainty to exhibit an artefact which stems from its practical implementation, and which we were not able to fully eliminate (see \autoref{fig:spatial_unc}). We therefore select \m{TTA+Ens} as the best-performing UQ method for predictive uncertainty and subsequent analysis. Notably, \m{TTA+Ens} recovers both meaningful and calibrated uncertainty estimates. Given that the traffic speeds can take values anywhere in $[0,255]$, an MPIW of at most $\sim23$ for a $90\%$ marginal coverage guarantee can be considered remarkably sharp, and ENCE $<1$ as well as $\rho_{sp} > 0.7$ show that the uncertainty is well-calibrated. The method thus successfully transfers across both spatial and temporal dimensions. The obtained methods comparison based on \autoref{tab:uq_comparison} is consistent with results for Bangkok and Barcelona, and when using a U-Net instead of U-Net++ model (see \autoref{app:results_table}).

There are two more observations to note. First, masking for traffic activity (\m{zero-mask}) results in a notable change in score values, although the relative ranking of UQ methods remains the same. Second, model performance both for mean estimation (consider the MSE scores) and uncertainty estimation is worse for Moscow than for Antwerp, even though it does not require an additional spatial transfer. In the case of no masking, these effects can be attributed to data sparsity, since $\sim70\%$ of cells for Antwerp (similarly for Bangkok and Barcelona) can be considered non-active, in contrast to $\sim39\%$ for Moscow. In the case of masking, this effect may be due to intrinsic differences in traffic dynamics, given that Moscow is a much larger city with higher traffic and road density, and thus potentially more complex dynamics. 

\subsection{Spatial distribution of uncertainty}
\label{subsec:uq-experiments-spatial}

\begin{figure}[t]
    \centering
    \includegraphics[width=\textwidth]{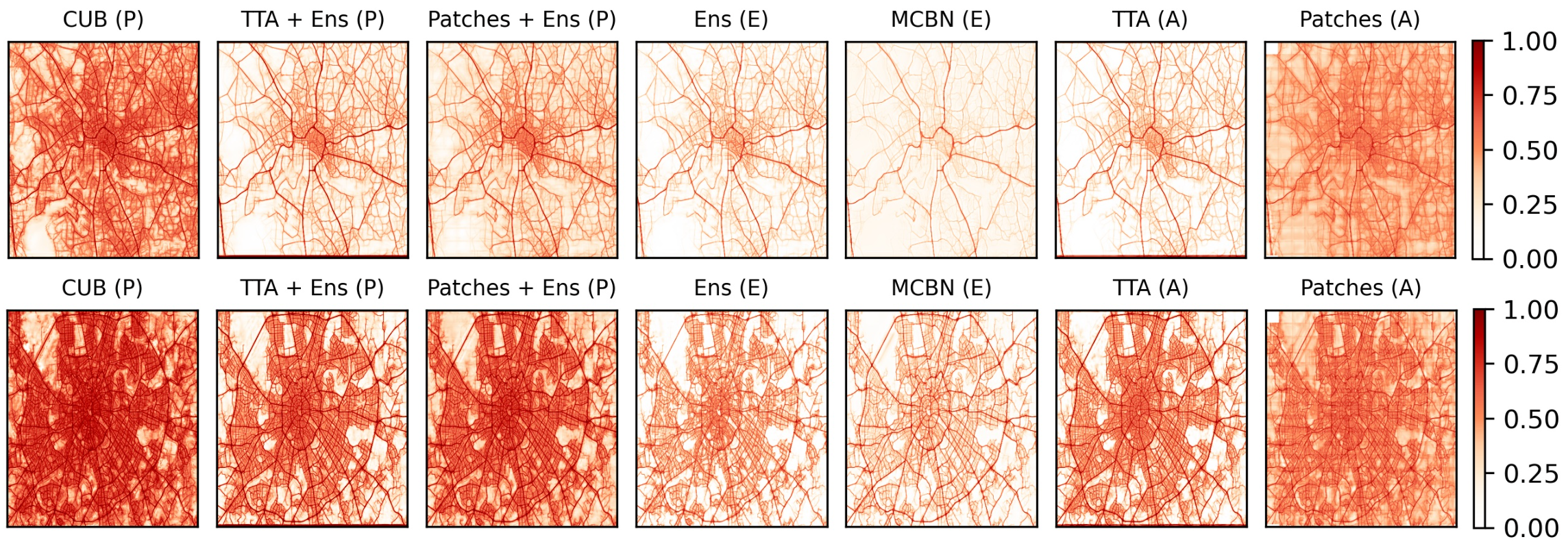}
    \caption{Spatial distribution of uncertainty estimates for Antwerp (\m{top}) and Moscow (\m{bottom}) across UQ methods. Values are logarithmised and normalized to $[0,1]$. Uncertainty values spatially recover the city's road network.
    }
    \label{fig:spatial_unc}
\end{figure}

\begin{figure}[t]
  \includegraphics[width=\columnwidth]{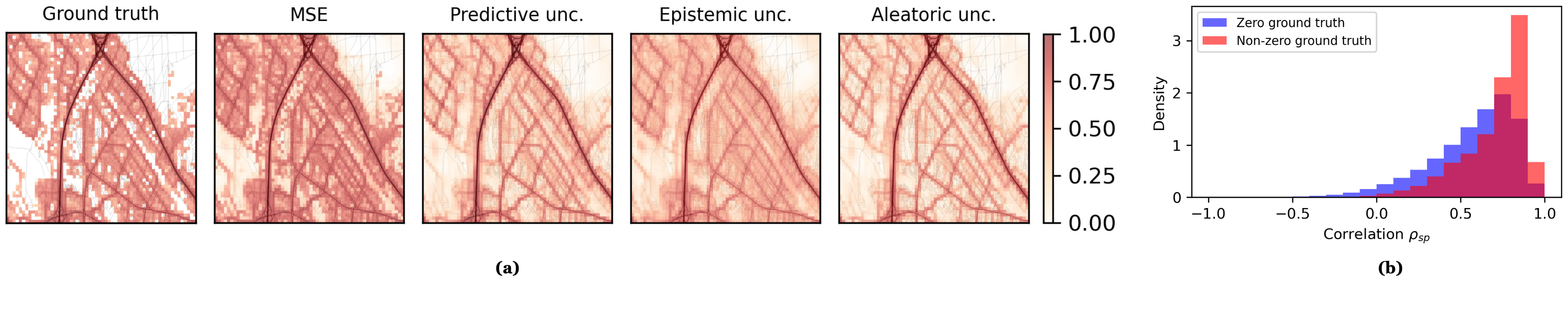}
  \caption{\m{Panel (a)}: Decomposed uncertainty values of \m{TTA+Ens} for a spatial crop in Moscow. Spatial alignment with both MSE and ground truth values is apparent. \m{Panel (b)}: Histogram of rank correlation $\rho_{sp}$ by ground truth mask type for Moscow. Both are strongly positive, and the correlation for non-zero ground truths is in fact higher.}
  \label{fig:spatial_unc_crop}
\end{figure}

We visualize the spatial distribution of uncertainty estimates across all UQ methods for Antwerp and Moscow in \autoref{fig:spatial_unc}. We first observe that uncertainty values spatially reflect the city's road network and to a varying degree across UQ methods. 
Note that \m{CUB} shows less spatial variation, 
while \m{Patches} exhibits a faint grid-like pattern that is an artefact from its sampling procedure (see also \autoref{app:patches}). Importantly, the magnitude of uncertainty values reflects the hierarchy of importance of a city's road network to its general traffic state, with uncertainty highest on major expressways and arterial roads \footnote{This results in a similar spatial pattern for constructed prediction intervals.}. This aligns with the expectation that traffic speeds are harder to predict when traffic dynamics are more complex, with both speed magnitude and variability higher on such major thoroughfares. There are no substantial differences in the spatial distribution of uncertainties between epistemic and aleatoric UQ methods; however, combining them to reflect total predictive uncertainty via \m{TTA+Ens} provides a more complete picture, both in terms of recovering the full road network as well as road hierarchy. This is highlighted in \autoref{fig:spatial_unc_crop}, where we zoom in on a city crop for Moscow that combines both an expressway junction as well as smaller residential roads. In addition, the close alignment of uncertainty estimates with both prediction error and ground truth values is apparent, highlighting the method's high calibration quality. Indeed, it is encouraging to see that the correlation when masking for non-zero ground truth values is in fact \emph{higher} than for zero values (see \autoref{fig:spatial_unc_crop}b), suggesting that the model's estimates are especially well-calibrated where it matters. The observed trends are consistent across all four test cities and both model architectures, indicating the quantification of spatially meaningful uncertainty estimates.

\subsection{Temporal distribution of uncertainty}
\label{subsec:uq-experiments-temporal}

\begin{figure}[th]
    \centering
    \includegraphics[width=\columnwidth]{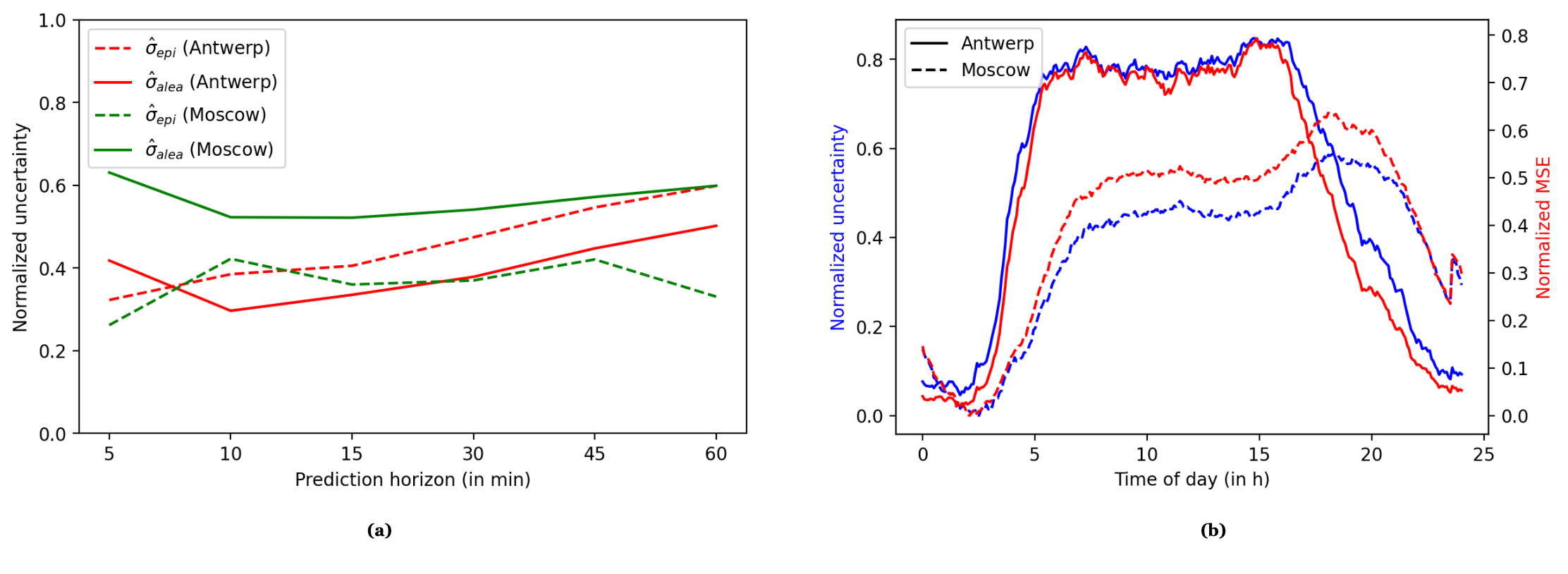}
    \caption{\m{Panel (a)}: Normalized aleatoric and epistemic uncertainties for Antwerp and Moscow against all six prediction horizons. The resulting curves are very stable across horizons. \m{Panel (b)}: Normalized uncertainty and MSE values for Antwerp and Moscow for an exemplary test day. Values are closely aligned also across the temporal dimension. All values obtained using \m{TTA+Ens}.}
    \label{fig:temporal_unc}
\end{figure}

Although our experiments primarily emphasize the 60 min ahead prediction horizon, we compare epistemic and aleatoric uncertainty estimates for a subset of test samples from Antwerp and Moscow across all six prediction horizons (i.e., 5, 10, 15, 30, 45 and 60 min ahead) in \autoref{fig:temporal_unc}. Intuitively, one might expect both an increase in predictive error and associated uncertainty as the prediction horizon increases. Although there may be a minor upward trend, we observe that the level of uncertainty remains relatively constant across all horizons. While perhaps somewhat surprising, this observation aligns with the findings of \cite{kreil2020surprising}, who report that the U-Net-based models exhibit a remarkably consistent performance on the \tfc dataset across prediction horizons. We also jointly display uncertainty and mean squared error values for a randomly selected test day (i.e., 288 coherent samples at 5 min sampling rate) for both cities. The temporal behaviour of uncertainty estimates aligns closely with that of traffic speed errors, which in turn behaves according to expected general traffic patterns. That is, errors strongly increase and decrease before morning and after evening rush hours when traffic variability increases respectively decreases. Predictive uncertainty follows the same pattern and is lowest during the night when traffic is infrequent. One can conclude that the alignment of predictive error and uncertainty is both spatially and temporally coherent. 

\section{Unsupervised outlier detection using epistemic uncertainty}
\label{sec:outlier-detection}

We now showcase how uncertainty quantification can be applied to outlier detection and interpreted with respect to city traffic dynamics. We do so by leveraging epistemic uncertainty estimates, which quantify the model’s uncertainty for unseen samples and are expected to be higher for unusual or anomalous observations \citep{kendall2017}. These estimates provide a signal to identify potential outliers in an exploratory unsupervised outlier setting. Obtained outliers are analysed temporally and spatially and related to city land use, indeed revealing links between quantified uncertainty and changes in traffic dynamics caused by different factors.




\subsection{Experimental setup}
\label{subsec:outliers-exp-setup}

\begin{algorithm}
\caption{Unsupervised outlier detection using epistemic uncertainty}
\label{alg:outlier_detection}
\begin{algorithmic}[1]
    \State \textbf{Input:} train data $\mathcal{D} = \{ (x_i, y_i) \}_{i=1}^{N}$, test data $\mathcal{D}^* = \{ (x^*_t, y^*_t) \}_{t=1}^{T}$, model $\hat{f}$, fixed time index $\tau$, outlier bound $\epsilon$
    
    \State \textbf{Output:} Binary outlier labels for $\mathcal{D}^*$ at time index $\tau$ per channel group (volume/speed) and per pixel.
    
    \State \textbf{Procedure:} 
    
    \State \quad $\hat{\sigma}_{1:N} \gets$ compute epistemic uncertainty estimates of $\hat{f}$ for $\mathcal{D}$ at time index $\tau$. \hfill \emph{\# reference values}
    
    \State \quad $\hat{\sigma}^{*}_{1:T} \gets$ compute epistemic uncertainty estimates of $\hat{f}$ for $\mathcal{D}^*$ at time index $\tau$. \hfill \emph{\# outlier candidates}

    \State \quad $\hat{g}(x) \gets$ estimate a density over $\hat{\sigma}_{1:N}$ using a non-parametric kernel density estimator, e.g. with Gaussian kernel.

    \State \quad \textbf{for} $t=1,\dots,T$ \textbf{do}

    \State \quad \quad Compute $\m{p-val}_{t} = \mathbb{P}(\hat{g}(x) \ge \hat{\sigma}^{*}_t \mid H_0)$, where $H_0$ is the null hypothesis that $\hat{\sigma}^{*}_t \sim \hat{g}(x)$. \hfill \emph{\# p-value per cell}

    \State \quad \quad Aggregate $\m{p-val}_{t}$ for volume channels using an aggregation scheme such as Fisher's method \citep{osborne1992fisher}: 
    
    $\m{p-val}_{t}^{vol} = \mathbb{P}(\mathcal{X}^2_{2k} \ge X^2 \mid H_0)$, where we have test statistic $X^2 = -2\sum_{j=1}^{k=4} \log(\m{p-val}_{t}^{j})$ and $H_0$ is the 
    
    null hypothesis that $X^2 \sim \mathcal{X}^2_{2k}$. $j=1,\dots,4$ are the directional channels. \hfill \emph{\# p-value for volume}

    \State \quad \quad Aggregate $\m{p-val}_{t}$ for speed channels in the same manner to obtain $\m{p-val}_{t}^{speed}$. \hfill \emph{\# p-value for speed}
    
    \State \quad \quad Set binary outlier label for volume channels as $\m{out-vol}_{t}$ = \emph{True} \textbf{if} $\m{p-val}_{t}^{vol} \ge \epsilon$ \textbf{else} \emph{False}.

    \State \quad \quad Set binary outlier label for speed channels as $\m{out-speed}_{t}$ = \emph{True} \textbf{if} $\m{p-val}_{t}^{speed} \ge \epsilon$ \textbf{else} \emph{False}.

    \State \quad \quad Set binary outlier label per pixel as $\m{out}_{t}$ = \emph{True} \textbf{if} ($\m{out-vol}_{t}$ = \emph{True} \textbf{or} $\m{out-speed}_{t}$ = \emph{True}) \textbf{else} \emph{False}.
    
    \State \quad \quad \textbf{return} ($\m{out-vol}_{t}$, $\m{out-speed}_{t}$, $\m{out}_{t}$) \hfill \emph{\# collect for each test sample}

    \State \quad \textbf{end for}
    
    \State \textbf{end procedure}
\end{algorithmic}
\end{algorithm}

Since the \tfc dataset lacks any ground truth outlier labels, we design a scenario in which we would expect changes in traffic dynamics to occur. We then analyse if identified outliers, as recovered by our unsupervised outlier detection approach, capture some (interpretable) notion of these changes. For example, we may want to discover that a specific city area records high outlier counts on specific weekdays and relate this to its land use, or how the onslaught of \covid~temporally affects traffic patterns. We create such an outlier setting by considering specific fixed time slots across test days, namely morning and evening rush hours (i.e., to and off-work peak travel times). These are especially interesting because they are the times of day at which traffic volume is typically highest, and thus provide a particularly condensed representation of the city's traffic state at a given moment. Additionally, they capture traffic patterns which are expected to be more sensitive to changes, e.g. when comparing traffic on weekdays and weekends or the impact of \covid~related work policies. In the remainder of this section, we mainly focus on morning rush hour but find general outlier detection patterns to translate to evening rush hour. Our unsupervised outlier detection framework is based on hypothesis testing and requires both training and test data for a given city. We therefore limit our analysis to test cities framed within the context of temporal transfer, namely Bangkok, Barcelona and Moscow. 
The selected time indices for morning rush hour based on observed traffic activity levels are fixed at 10:00 AM for Bangkok, 08:20 AM for Barcelona, and 09:20 AM for Moscow. Since our test data stretches 90-day time periods and we fix a constant time slot per day, we evaluate 90 temporally coherent test samples for potential outliers \emph{per cell} \footnote{For a given pixel $(i, j)$ and channel $c$, the tuple $(i, j, c)$ constitutes a cell.}, which can represent both temporal and spatial traffic changes.

To detect outliers (i.e., anomalous observations) in an unsupervised manner, we employ statistical hypothesis testing by comparing test uncertainties against the distribution of training uncertainties obtained using a kernel density estimator. The ideas behind our approach are well-established in the outlier detection literature (see, e.g., \cite{beckman1983outlier}), make relatively few assumptions, and can be easily applied \emph{post-hoc} to any model outputs. Specifically, our approach, as outlined in Algorithm \ref{alg:outlier_detection}, makes two main assumptions.

\emph{Assumption 1: Anomalous observations occur in low-probability regions of the data-generating process, i.e., they are far less frequent than non-anomalous observations.} This is a standard assumption for statistical outlier detection \citep{chandola2009}, and it allows us to compare potential outliers to the density estimate of samples seen by the model. Instances with a low probability of occurrence, expressed by high epistemic uncertainty on their prediction, can then be identified based on hypothesis testing.

\emph{Assumption 2: The available training data from the data-generating process is a representative sample that contains none or only very few outliers.} If this were not the case, then obtained density estimates for uncertainty values would not be representative of the non-anomalous case, and we would fail to recover significant outliers via testing. We observe in practice that results suggest support for the validity of the assumption (see, e.g., \autoref{fig:out-hist-moscow}). 

Regarding experimental design, we rely on deep ensembles (\m{Ens}) to obtain epistemic uncertainty estimates, and use a Gaussian kernel for kernel density estimation. While we experiment with other kernel options, we find a Gaussian kernel to provide the best fit. Although working directly with the empirical histograms is possible, a smooth density approximation circumvents difficult choices such as optimal bin size and permits more accurate testing. Finally, we fix the outlier bound $\epsilon$ at 0.001, i.e., any p-values below $0.1\%$ probability mass will classify as outliers. While seemingly restrictive at first, we find outliers to be quite robust w.r.t. the choice of $\epsilon$, meaning their associated p-values are small enough to identify as outliers irrespective of a reasonable choice for $\epsilon$ (see \autoref{app:outliers}). A tighter outlier threshold may help shield against false positives to some extent.

\subsection{General observations}
\label{subsec:outliers-general}

\begin{figure}[t]
  \centering
  \begin{subfigure}[b]{0.65\textwidth}
    \includegraphics[width=\linewidth]{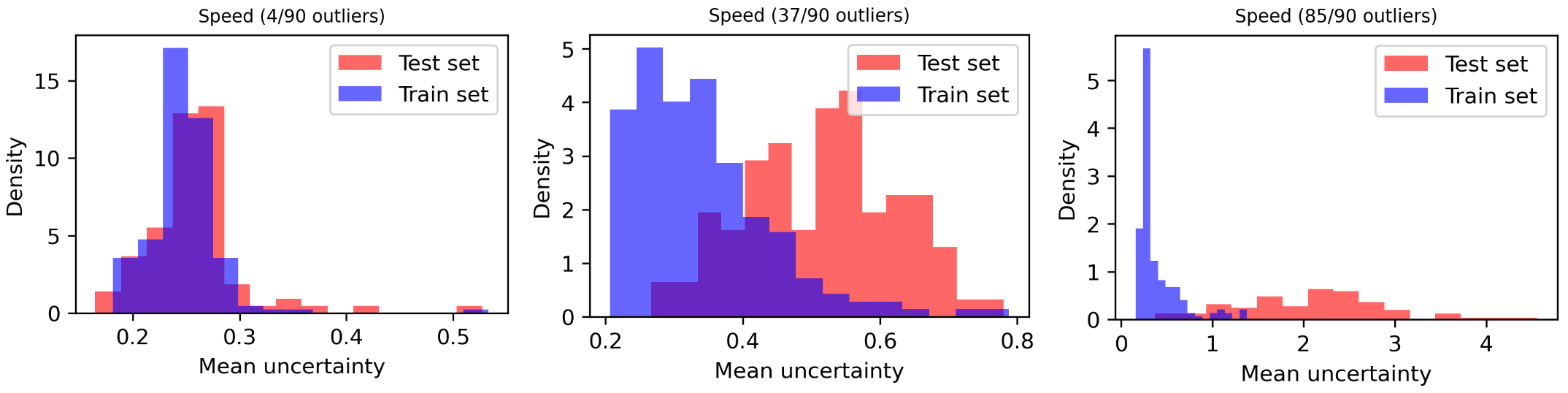}
    \caption{}
    \label{fig:out-hist-moscow}
  \end{subfigure}
  \hfill
  \begin{subfigure}[b]{0.3\textwidth}
    \centering
    \begin{tabular}{l r r r}
        \toprule
        \textbf{Outliers [\%]} & \textbf{Vol.} & \textbf{Speed} & \textbf{Pixel} \\
        \midrule
        Bangkok & 0.37 & 0.62 & 0.84 \\
        Barcelona & 0.45 & 0.64 & 0.89 \\
        Moscow & 4.98 &	5.25 & 8.01 \\
        \bottomrule
    \end{tabular}
    \vspace{4mm}
    \caption{}
    \label{tab:results-out-stats}
  \end{subfigure}
  \caption{\m{Panel (a)}: Histograms of uncertainty estimates for three exemplary pixels from Moscow. We observe how the magnitude of deviation between train and test uncertainties directly translates to recovered pixel outlier counts. \m{Panel (b)}: Counts of recovered outlier labels as a percentage of total pixels ($495 \times 436$). Moscow records notably higher outlier counts relative to the other two cities for the same fixed outlier bound $\epsilon = 0.001$.}
  \label{fig:results-out-stats}
\end{figure}

We quantify the number of recovered outlier labels following Algorithm \ref{alg:outlier_detection} as a fraction of total pixels for all three test sets in \autoref{tab:results-out-stats}. In comparison to Bangkok and Barcelona, Moscow records notably higher fractions of outliers, pointing to a higher model uncertainty for the given test data. Given that all three cities experience a temporal shift between train and test data, this suggests that traffic dynamics in Moscow have experienced a particularly strong change. Naturally, the difference in data sparsity ($\sim70\%$ sparse cells for Bangkok and Barcelona vs. $\sim39\%$ for Moscow) and choice of outlier bound play a strong role in the magnitude of recovered counts. However, we find their relative difference to prevail across the selection of $\epsilon$. 
An intuitive visualisation on the occurrence of positive outlier labels is given in \autoref{fig:out-hist-moscow}, wherein we contrast the empirical distributions of train and test uncertainty estimates for exemplary pixels. A stronger change in traffic dynamics - as presumably caused by temporal shift - results in higher model surprisal as measured by epistemic uncertainty, and is flagged by our unsupervised detection method as potentially more anomalous behaviour. This signalling is similar to e.g. \cite{lakshminarayanan2017} who identify out-of-distribution data via shifts in measured entropy capturing higher model surprisal.

\subsection{Temporal outlier patterns}
\label{subsec:outliers-temporal}

\begin{figure}[ht]
	\includegraphics[width=\textwidth]{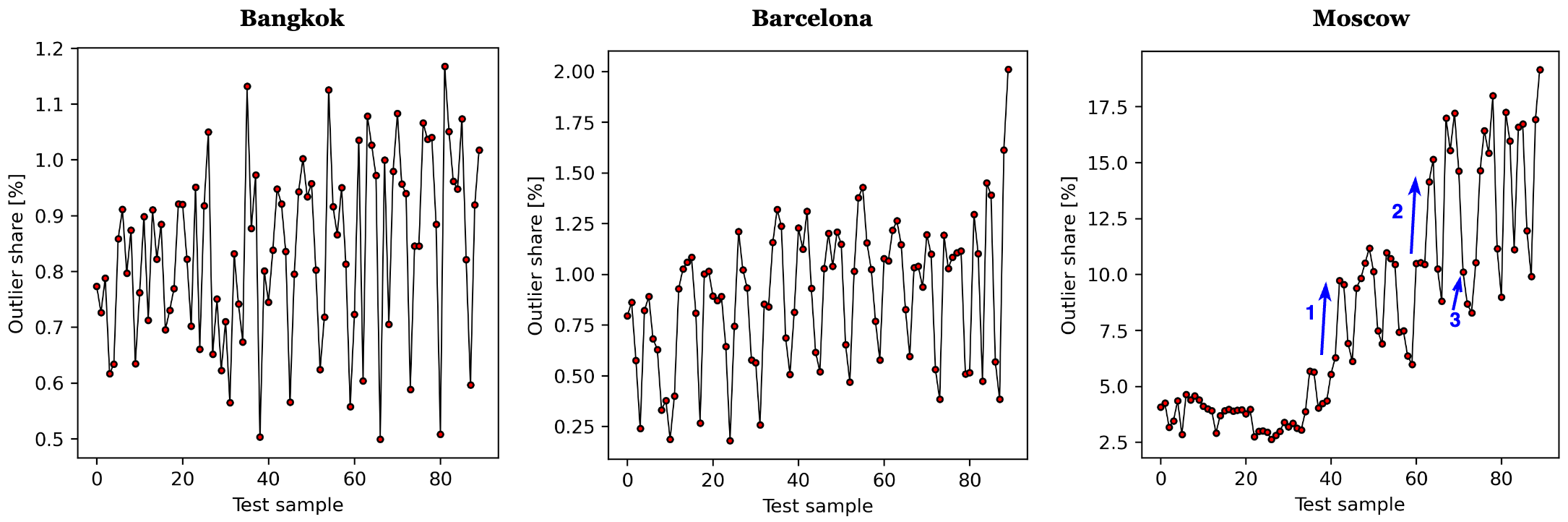}
	\centering
	\caption{Outlier counts per city visualised across test samples. Outlier share are pixel outliers as a fraction of total pixels ($495 \times 436$). 
    Specific occurrences marked by blue arrows for Moscow are explained in the text.}
	\label{fig:out-temporal}
\end{figure} 

We are now specifically interested in viewing changes in the magnitude of outlier counts across time as an indicator for shifts in underlying traffic dynamics. We therefore visualise recovered pixel outliers across our set of temporally coherent test samples spanning the time period from 02/04/2020 to 30/06/2020 in \autoref{fig:out-temporal}. Note that each test sample corresponds to city-wide morning rush hour on a specific day from that period. We immediately observe that there is some variation in outliers across different days. We also observe a positive trend in outlier counts for Moscow, while outlier shares for Bangkok and Barcelona remain fairly stable. Given that our test data is fully within a \covid~time period while our train data is not, we might expect changes in traffic dynamics - as expressed by recovered outlier shares - to coincide with the effect of \covid~regulations. Similarly, we might observe an impact of the day of the week or special occasions such as public holidays on the city's traffic state. Naturally, there are a series of other possible factors contributing to any observed temporal shift. However, for lack of outlier ground truth labels (our approach is \emph{unsupervised}) any accurate disentanglement of contributing aspects is practically impossible to obtain. We therefore focus on the above factors and qualitatively relate each city in more detail to them via \autoref{fig:out-temporal}. 

\m{Bangkok.} We do not observe any distinct pattern in obtained outlier shares for Bangkok. As with most of Thailand in general, Bangkok remained in a fairly strict state of lockdown during the time period covered by our test data, including bans on inter-provincial travel \citep{bangkokcovid, bangkokcovid2}. Some slight easings, such as a lift on said travel ban on June 1st, coincide with spikes in outlier counts. 
Since recovered outlier counts are very low, it is likely that spatial effects, i.e., generally difficult traffic conditions in a given area, contribute a larger share than temporal effects.

\m{Barcelona.} Similar to Bangkok we do not observe a distinct general trend, albeit perhaps a slight increase towards the latter part of the test set. Barcelona remained under strict lockdown conditions for most of the time period, with a state of emergency being repeatedly extended \citep{barcelonacovid}. A lift on most mobility restrictions and the re-opening of EU borders on June 30th coincides with a sharp rise in outlier counts at the end of the time period, indicating a change in traffic activity. We also identify a distinct pattern in traffic activity between weekdays and weekends, with weekends having much lower outlier counts. The regularity of the pattern improves towards the latter half of the time period when some minor lockdown easings seemingly induce more structured traffic activity. Given the generally low outlier counts, likewise, a spatially-related effect outweighing temporal effects may be possible. For example, we observe that one spatial region exhibiting consistently high outlier counts for Barcelona is an industrial business park (see \autoref{app:outliers}).

\m{Moscow.} In contrast to Bangkok and Barcelona, Moscow exhibits a very distinct upwards trend in outlier counts across the time period. While initially a strict nationwide lockdown was enforced affecting all sectors \citep{moscowcovid}, an initial lockdown easing on May 11th is expressed by a general increase in outlier counts, suggesting a notable change in traffic dynamics (\m{blue arrow \#1}). An additional easing of lockdown measures on June 9-10th leads to a further change in general traffic activity as more people leave the confinements of their homes on a regular basis (\m{blue arrow \#2}). In addition, the distinction between weekends and weekdays becomes more pronounced. We are also able to identify individual deviations such as a drop in outliers on Friday, June 12th, typically a regular working day but that year coincides with a public holiday (\m{blue arrow \#3}). Given the higher percentage of outlier counts and their occurrence in areas across the whole city map (see next section), it is reasonable to claim that a temporal effect on city-wide traffic dynamics can be observed. 

\subsection{Spatial outlier patterns: a case study for Moscow}
\label{subsec:outliers-spatial}

\begin{figure}[ht]
	\includegraphics[width=\textwidth]{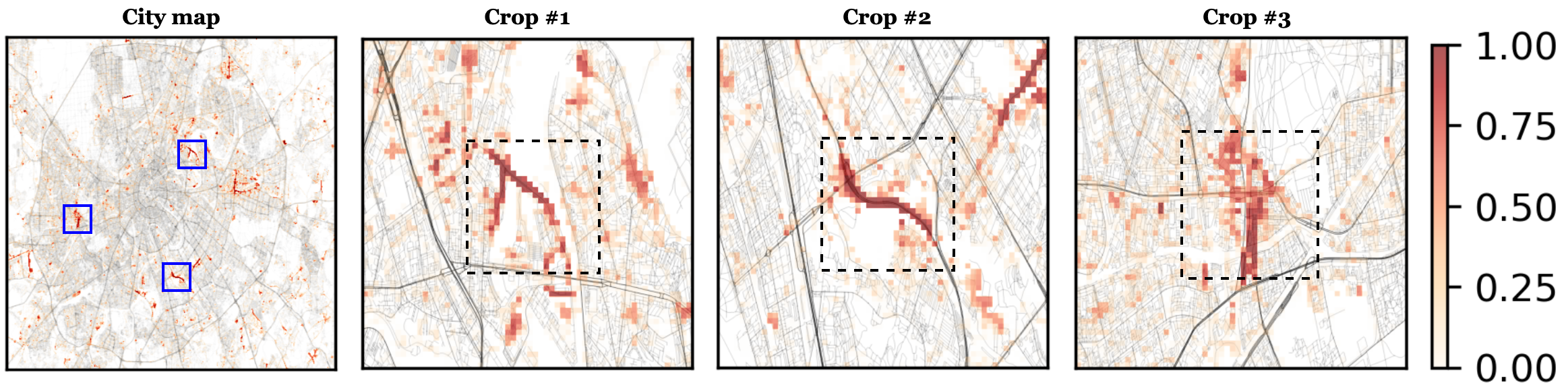}
	\centering
	\caption{Spatial distribution of pixel outlier shares for Moscow, both on the whole city map and three selected city crops (\m{blue squares}). For each city crop, we visualize the central spatial area (\m{black squares}). Outlier shares are computed as a percentage of total time steps, i.e., a max. of 90 outliers is possible. Outliers occur across the whole city but are spatially clustered along the road network.}
	\label{fig:out-spatial}
\end{figure} 

We now aim to reason about identified outlier regions in more detail by considering both temporal and local spatial traffic behaviour that may induce higher model uncertainty. 
Specifically, there is an obvious temporal pattern to be expected for any ground truth traffic state (i.e., volume and speed) for morning rush hour: we expect lower traffic volume on weekends than on weekdays, and potentially (but not necessarily) higher traffic speeds as a side-effect. In regards to spatial patterns, we consider a series of image crops of about $30 \times 30$ pixels that match city regions recording a high outlier density, corresponding to landscape sections of $\sim3\text{km}^2$ (see \autoref{fig:out-spatial}). The \emph{central} area of each crop, containing the bulk of outliers ($\sim1\text{km}^2$), is then contrasted against the surrounding city (the \emph{context} area). We inspect traffic volume and speed values across both temporal (weekend vs. weekday) and spatial (central vs. context area) partitions for train as well as test data. Any observed mismatches across either partitions or data sets should ideally provide some insight into what contributes to higher epistemic uncertainty and thus positive outlier flags. We then aim to further corroborate the occurrence of outlier labels by looking at the city area's land use. We run our analysis as a case study for three city regions of high outlier density in Moscow, which itself recorded the largest outlier share among the cities considered in \autoref{tab:results-out-stats}. 
The spatial distribution of outlier shares across test samples for both the full city and selected crops is visualized in \autoref{fig:out-spatial}. We observe that instances of outliers occur across the whole city and are aligned with the road network, and that regions of high outlier density are spatially clustered. We now examine three such regions \footnote{Approximate GPS locations for the three city crops are (left to right): (\m{55°41'31.5" N, 37°40'45.8" E}), (\m{55°42'44.4" N, 37°29'37.9" E}) and (\m{55°50'50.0" N 37°35'06.4" E}).} in greater detail following the strategy outlined above. The exact figures for our interpretation can be found in \autoref{app:outliers}.

\begin{figure}[!t]
	\includegraphics[width=\textwidth]{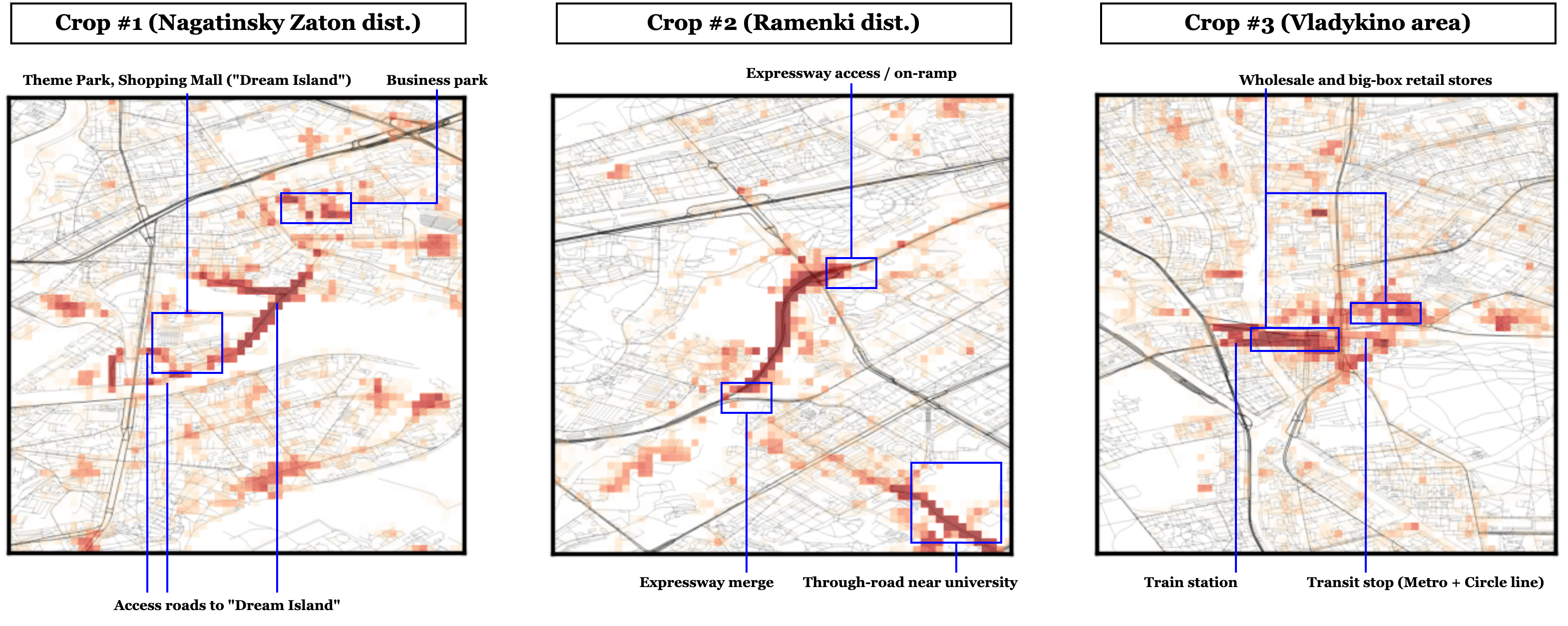}
	\centering
	\caption{Identification and annotation of land use for key features of the Moscow city crops selected in \autoref{fig:out-spatial}. The city crops are flipped by 90° to be geographically north-aligned. Details on their interpretation are found in the text.}
	\label{fig:out-landuse}
\end{figure}

\m{Crop \#1}. For this city crop in the Nagatinsky Zaton district of Moscow, we observe that the expected temporal pattern (weekend vs. weekday) is apparent on the test set, but does not occur on the train set for the crop's central area. It occurs for both sets for the crop's context area. We interpret these outlier instances to be primarily caused by a temporal mismatch, in that the model is faced with an unexpected regularity pattern at test time that is not exhibited at train time, resulting in higher uncertainty. This interpretation aligns with the crop's land use in \autoref{fig:out-landuse}, which contains a centrally located theme park with shopping and attractions (`Dream Island’). Most of the outliers are observed on the access roads leading to the theme park. One would expect that in pre-\covid~times such a theme park would attract visitors irrespective of the specific day, recording high traffic activity throughout. However, during \covid ~the theme park was forced to remain closed, resulting in a more regular temporal pattern matching its spatial context, since there is no additional traffic frequenting the area for the purpose of visiting the park. Thus a temporal mismatch caused by the impact of \covid~on the test data is a possible factor for arising uncertainty and subsequent outlier flags.

\m{Crop \#2}. For this city crop in the Ramenki district of Moscow, we observe that the temporal pattern tends to rather occur on the test data, but not train data for both central and context areas of the crop. Similarly to the previous crop, one can conclude a temporal mismatch in terms of regularity pattern. Furthermore, this crop exhibits high traffic activity and irregularity in pre-\covid~times, which suggests intrinsically difficult traffic dynamics need to be captured by the model, resulting in generally higher uncertainty. This is supported by \autoref{fig:out-landuse}, since we note this city section contains a centrally located city expressway, with both a complex expressway on-ramp combining heterogeneous traffic flows (via a join of multiple roads at different hierarchy levels) and an expressway merge. In that sense, the positive outlier flags can be considered a signal to warrant closer inspection of this expressway section and highlight the need for traffic design improvements.

\m{Crop \#3}. For this city crop in the Vladykino area of Moscow, we observe a similar temporal and spatial pattern behaviour as for \m{Crop \#2} and make an analogous conclusion with regard to arising outliers. In this case, the intrinsically complex traffic dynamics are in one part due to the area being a hub of different transportation modalities, combining train, metro and Circle line stations. We also expect heterogeneous traffic flows due to the high density of wholesale and big-box retail stores, resulting in e.g. many parking entrances and exits, cargo on- and off-loading and a mix of personal and commercial vehicles. Once more, the positive outlier flags can be considered an indicator of a generally difficult traffic situation.

Our analysis in this case study highlights that identified outliers can be reasonably related to city land use and may provide an indicator for situations exhibiting difficult traffic behaviour; and that both temporal and spatial traffic patterns can impact model uncertainty to various degrees and in different ways. Ultimately, this can be traced back to the inherently complex spatio-temporal nature of traffic data.

\section{Conclusion}
\label{sec:conclusion}

This work aims to showcase the usefulness of uncertainty-aware models for traffic prediction tasks, and presents general approaches that can enrich the current state of traffic analysis with uncertainty quantification. We theoretically motivated a series of \emph{post-hoc} applicable UQ methods and evaluated them on a comprehensive traffic dataset. Our results indicate that a proper combination of aleatoric and epistemic uncertainty quantification methods 
provides accurate uncertainty estimates and generalises well both across space (i.e., multiple cities) and time (i.e., different years). We demonstrated that obtained uncertainty estimates relate in multiple ways to the city's underlying road network and its traffic dynamics, and their usefulness in the context of an unsupervised outlier detection task was showcased. We additionally highlighted the utility of an image-based data representation for uncertainty-aware traffic prediction tasks, since it allows for a city-wide analysis and comparison of uncertainty estimates. The approaches we suggest are general, virtually distribution-free and readily applicable \emph{post-hoc}, allowing for a straightforward adaptation to different tasks and settings. 

There are several limitations and arising opportunities for future work regarding the experiments conducted in this study. Firstly, even though the \tfc dataset can be considered spatially dense relative to, e.g., loop counter data, we still face the difficulty of regions of spatial sparsity. Experimenting with different approaches to handling such regions, e.g. via masked training, may alleviate some associated challenges. Secondly, some of our experiments are computationally limited by the data's high dimensionality and large memory footprints. As a result, we cannot maximize the use of available calibration and test data for inference. Effectively scaling up to fully harness available data, e.g., having a larger calibration set to construct prediction intervals, may improve results and provide a more robust evaluation. Additionally, evaluating more UQ methods that perhaps impose mild distributional assumptions and experimenting with more model architectures or hyperparameter tuning may yield further improvements. Finally, we may want to experiment with additional outlier detection scenarios or strive to obtain ground truth outlier annotations. Resulting avenues of future work include addressing the above limitations; considering further applications of presented UQ methods in the context of traffic prediction, such as their incorporation into human-machine interaction systems \citep{wang2020visual} or extension to other suitable traffic datasets; and more closely exploring the conformal prediction framework, which can also be framed as a tool for outlier detection using obtained prediction intervals. Ultimately, this work aims to raise awareness of the benefits of uncertainty-aware deep learning models for traffic prediction, and showcase their potential value contribution to analysing and deploying intelligent transportation systems for our cities.

\section*{CRediT authorship contribution statement}
\textbf{Alexander Timans:} Methodology, Software, Visualization, Writing - Original Draft, Writing - Review \& Editing; \textbf{Nina Wiedemann:} Conceptualization, Software, Writing - Review \& Editing; \textbf{Nishant Kumar:} Conceptualization, Validation, Writing - Review \& Editing; \textbf{Ye Hong:} Conceptualization, Writing - Review \& Editing; \textbf{Martin Raubal:} Writing - Review \& Editing, Funding acquisition

\section*{Declaration of competing interest}
The authors declare that they have no known competing financial interests or personal relationships that could have appeared to influence the work reported in this paper.

\section*{Acknowledgments and funding}
We thank HERE Technologies and the IARAI for providing the \tfc dataset and the associated challenge. We thank Christian Eichenberger and Moritz Neun from IARAI for their comments on an early version of this work. This research was partially supported by The Hasler Foundation [Grant 1-008062]. Furthermore, NK from Future Resilient Systems project at the Singapore-ETH Centre (SEC) is supported by the National Research Foundation, Prime Minister's Office, Singapore under its Campus for Research Excellence and Technological Enterprise (CREATE) programme.

\bibliographystyle{abbrvnat}
\bibliography{references}

\newpage

\appendix
\section*{Appendices}

\section{Model training}
\label{app:model_training}

All performed experiments were implemented in \m{Python 3.8}, with primary use of the software packages \m{PyTorch} \citep{paszke2019pytorch} and \m{NumPy} \citep{harris2020numpy} for computations, and \m{Matplotlib} \citep{hunter2007matplotlib} for visualizations. A full list with all used packages and their versioning, as well as instructions to reproduce the software environment and sample run commands can be found in the accompanying code repository. Most computations, including model training and inference, were performed on standard cluster CPUs and two \m{NVIDIA TITAN RTX} GPUs. U-Net and U-Net++ predictive models are trained using their default architectures - aside from supplementing them with batch normalization layers - and default parameter settings, and trained with MSE loss functions and Adam optimizer. For fixed parameter values see also the code repository. We fix a random seed per model and train an ensemble of 5 models per architecture type with batch size 12, which was the largest batch size we could fit into memory. One epoch on the full training dataset took approx. $5-7$ GPU hours trained in parallel on two \m{NVIDIA TITAN RTX} GPUs. We use the best-performing U-Net resp. U-Net++ model by MSE validation loss for all UQ methods requiring a single base model, and all five for any method requiring deep ensembles. Our MSE validation losses stabilize at a value of $\sim50$, which is consistent with MSE scores observed in the \tfc challenge \citep{eichenberger2022traffic4cast, lu2021, wiedemann2021traffic}.

\section{Conformal prediction intervals}
\label{app:coverage}

\begin{figure}[h]
    \centering
    \includegraphics[width=0.6\columnwidth]{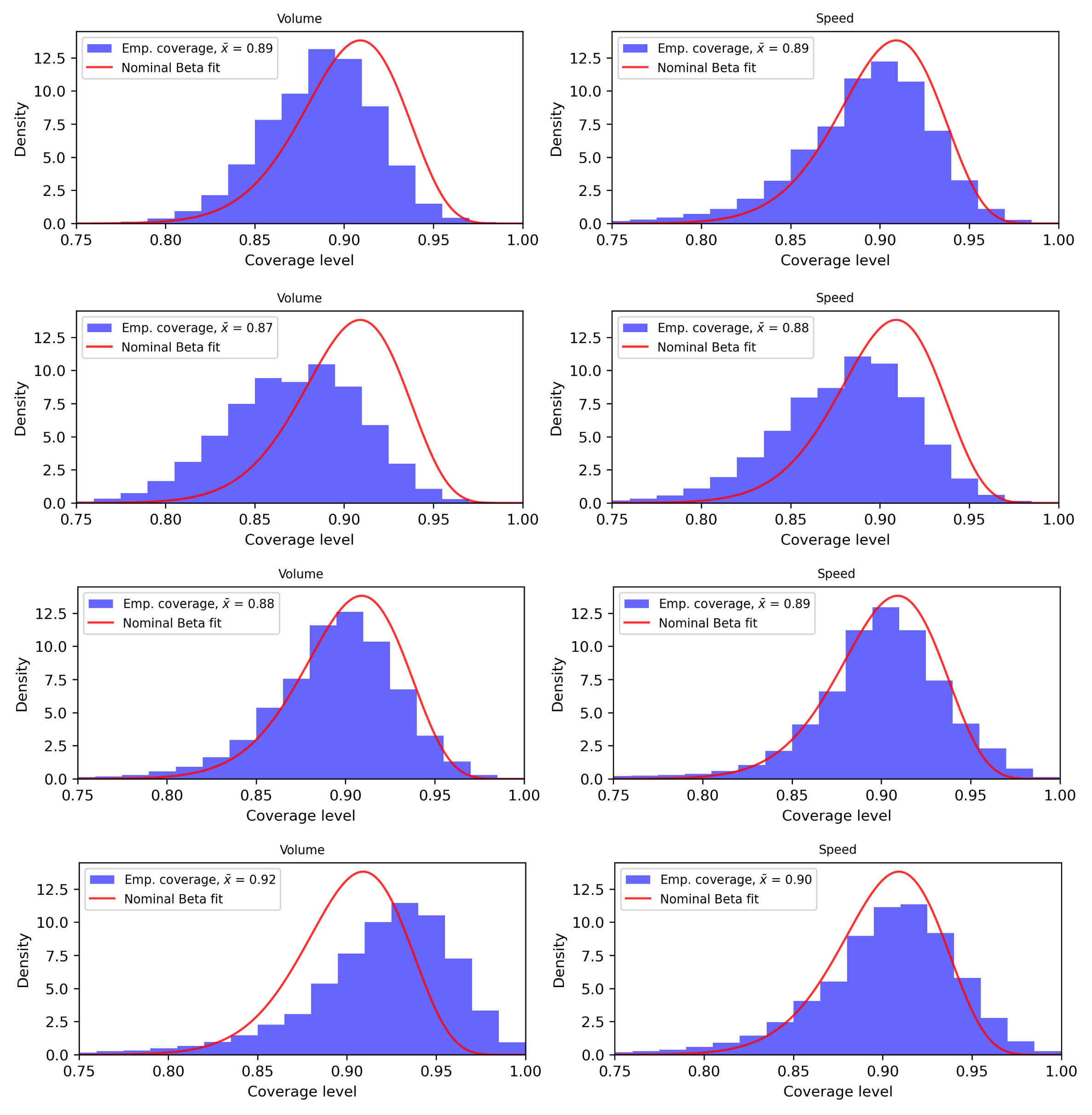}
    \caption{Empirical coverage histograms across test samples against nominal coverage \m{Beta} distribution fits for \m{TTA+Ens}. \m{Left:} volume channels, \m{right:} speed channels. \m{From top to bottom}: Antwerp, Bangkok, Barcelona, Moscow.}
    \label{fig:cp_cov}
\end{figure}

We construct conformal prediction intervals (PIs) following \cite{angelopoulos2023conformal}. For each sample $(x_i,y_i)$ in our calibration dataset $\mathcal{D}^{cal} = \{(x_i,y_i)\}_{i=1}^C$ we compute normalized conformity scores as $s(x_i,y_i) = |y_i-\hat{\mu}(x_i)|/\hat{\sigma}(x_i)$, and take a \emph{cell-level} quantile $\hat{q}$ as the $\lceil (1-\alpha)(C+1)\rceil/C$ quantile over scores for that particular cell. To validate proper implementation of conformal prediction, we compare the distribution of empirical coverages across test samples against the nominal coverage distribution, which has been shown to be a Beta distribution of shape $\m{Beta}(C+1-l,\,\,l),\,\,l=\lfloor (C+1)\,\alpha\rfloor$ \citep{vovk2012conditional}. That is, we compute the mean empirical coverage across cells and channel groups (volume/speed) per test sample and plot the histogram over test samples, separately for each of the four test cities and for each channel group. Given our choice of $\alpha=0.1$ and usage of $C=100$ calibration samples, we can expect empirical coverages of around $90 \pm 5\%$. We observe in \autoref{fig:cp_cov} that empirical coverages do not deviate drastically from the nominal fit, suggesting a valid implementation of conformal prediction intervals. While \citet{angelopoulos2023conformal} recommend calibration sets in the size of 1000 samples, this was not feasible due to memory limitations dictated by our large data shapes.

\section{Uncertainty quantification methods: further details}
\label{app:uq_implementation}

\subsection{Deep ensembles (\m{Ens})}
An arguably important component for treating ensemble member weights $\theta_m$ as \IID Monte Carlo samples is to induce sufficient ensemble diversity. Perhaps surprisingly, simple randomization schemes in the training process such as random weights initialization and random shuffling of the training data has been repeatedly found to induce sufficient diversity in the ensemble for empirically strong performance \citep{lakshminarayanan2017, lang2022, gustafsson2020a}. Recent work e.g. by \citet{abe2022} has indeed suggested that the importance of ensemble diversity for strong predictive performance is lower than expected.

\subsection{Monte Carlo batch normalization (\m{MCBN})}

For \m{MCBN}, we run $M=10$ stochastic forward passes through the model with sampled mini-batches from the training data of size 12, equivalent to that used at training time. It is important to note that the mini-batch used at test time should still be sampled from the available training data, i.e. $\mathcal{B} \sim X$, and the batch size $|\mathcal{B}|$ should be kept consistent with that used at train time. This is to ensure that the distribution from which $\omega$ is sampled at test time is indeed equivalent to the approximate posterior $q_{\theta'}(\omega)$ optimized during training \citep{teye2018bayesian}. To formally obtain \IID Monte Carlo samples, the training samples used for computing $\omega$ should be drawn individually with replacement, while in practice one samples a whole mini-batch at once. However, for sufficient training steps \citet{teye2018bayesian} show that the distribution of sampled batch members converges to the case under \IID assumptions. Note that our final model only induces stochasticity in the first two BN layers of the model, following observations made by \cite{teye2018bayesian} (see section 4.4 in their paper). While we also experiment with all BN layers being stochastic, we did not find a better quality in resulting uncertainty estimates at the cost of substantially longer inference times.

\subsection{Test time augmentation (\m{TTA})}

For \m{TTA}, images are padded to quadratic size of $496 \times 496$ and the following augmentations are applied: vertical \& horizontal flips, rotations in \{90\textdegree, 180\textdegree, 270\textdegree\} and vertical flips with \{90\textdegree, 270\textdegree\} rotation, resulting in a total of seven spatially reversible augmentations. As mentioned previously, pixel brightness encodes traffic information and we therefore do not employ e.g. saturation changes that will perturb the spatial traffic dimension. Naturally, we expect the uncertainty estimation quality to improve with larger sample counts. However, obtaining more transformations while maintaining \IID assumptions can prove a practical hurdle for image-to-image tasks, as there may only be a finite set of reversible transformations and combinations thereof applicable to the data, as in our case. Finally, in practice we do not take our point estimate $\hat{\mu}$ as the mean over all augmented predictions, but rather as the prediction on the original (unaugmented) image. This was mainly due to observing that it provides a slight improvement in MSE score over averaging. 

\subsection{Patch-based uncertainty (\m{Patches})}

It is important to observe that following the approach outlined in \autoref{subsec:uq-methods-patches} on a pixel-by-pixel basis generates large computational overhead. This occurs since we discard predictions for the non-central pixels within a given context patch (i.e. those which are not $\mathcal{P}$), and which have to be subsequently re-evaluated when it is that pixel’s `turn' for uncertainty estimation. We attempt to alleviate this by extracting context patches at test time in a more systematic sliding-window manner and predicting for multiple pixels simultaneously, as well as introducing a stride $s$ for the window to reduce the collection of highly correlated patches that may contain low informational value for UQ. See \autoref{app:patches} below for more details.

\subsection{Predictive uncertainty (\m{TTA+Ens})}

\begin{figure}[h]
    \centering
    \includegraphics[width=0.5\columnwidth]{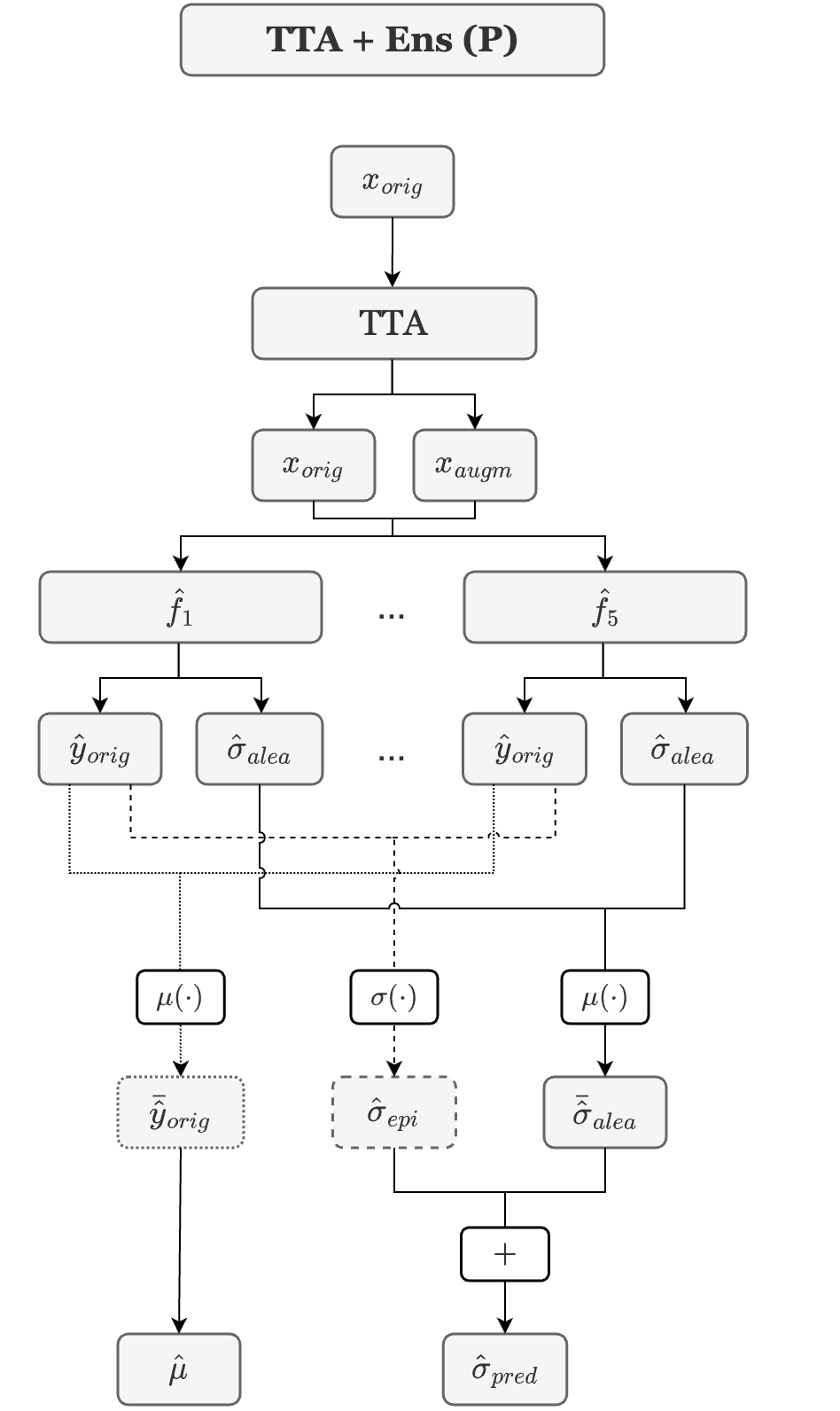}
    \caption{Schematic of predictive uncertainty method \m{TTA+Ens}, which combines test time augmentation for aleatoric and deep ensembling for epistemic uncertainty estimation.}
    \label{fig:tta_ens}
\end{figure}

We visualize the practical implementation of our predictive uncertainty method \m{TTA+Ens} in \autoref{fig:tta_ens}. For a given test sample $x_{orig}$ we first apply the data augmentations to generate augmented versions $x_{augm}$ of the sample. We subsequently feed both the original and augmented versions into the model ensemble. For each model we recover a point prediction $\hat{y}_{orig}$ on the unaugmented sample and a measure of aleatoric uncertainty $\hat{\sigma}_{alea}$ as the variation in predictions across both the original and augmented versions of the test sample. We then take the mean over point estimates across all ensemble members $\bar{\hat{y}}_{orig}$ as a final point estimate $\hat{\mu}$. We take the variation over predictions on the original (unaugmented) sample across ensemble members as a measure of epistemic uncertainty $\hat{\sigma}_{epi}$. And we take the mean over aleatoric uncertainty estimates across ensemble members as an `averaged’ aleatoric uncertainty estimate $\bar{\hat{\sigma}}_{alea}$. Lastly, we combine $\hat{\sigma}_{epi}$ and $\bar{\hat{\sigma}}_{alea}$ additively following \autoref{subsec:uq-methods-predictive} to obtain a final predictive uncertainty estimate $\hat{\sigma}_{pred}$.

\section{Patch-based uncertainty: further analysis}
\label{app:patches}

We do not select \m{Patches+Ens} as our final approach for modelling predictive uncertainty (instead we opt for \m{TTA+Ens}) because \m{Patches} exhibits a grid-like artefact with regards to uncertainty estimates, as observed in \autoref{fig:spatial_unc}. Specifically, there is an increase in uncertainty around the borders of each sampled patch - as determined by the choice of stride $s$ and patch size $d$ - that is not rooted in ground truth traffic information, but originates from our modelling procedure. Despite efforts made, we were not able to properly eliminate this grid-like artefact, and believe it to stem from our sampling procedure using a sliding window approach following \citet{wiedemann2021traffic}. Note that they do not observe such an effect when employing a patch-based approach merely for improving predictive performance, but we observe it to arise for uncertainty estimation. Despite this modelling artefact, \m{Patches+Ens} shows remarkably strong performance in \autoref{tab:uq_comparison}, suggesting that it may be worth considering alternative sampling procedures, and that patch-based procedures should be further explored as a viable approach for \emph{post-hoc} aleatoric UQ. In the following we include more details on \m{Patches} that may provide impulses for such exploration, and be of interest to general research in UQ. 

\subsection{Computational complexity}
\label{app:patches_complexity}

Our goal is to obtain an aleatoric uncertainty estimate for each pixel of the gridded input image. Let $N$ be the number of context patches $C$ that must be generated to obtain such an estimate $\sigma_{alea}$ for all pixels. $N$ dictates the computational complexity of the method and as such the required inference time. Since each pixel can be covered by at most $d^2$ context patches of size $d \times d$, i.e. $M \leq d^2$, considering each pixel in an image of size $I_h \times I_w$ independently for uncertainty estimation necessitates $N = M \cdot I_h \cdot I_w \in O(d^2 \cdot I_h \cdot I_w)$ patch predictions, which is impractical. However, $N$ can be greatly reduced by exploiting that each model pass provides predictions for all $d^2$ pixels per context patch. Instead of sampling context patches for each pixel separately, these can instead be systematically extracted using a sliding window over the image. In order to obtain the maximal Monte Carlo sample counts per pixel $M=d^2$ we require predictions across all possible context patches covering the image, i.e. $N = I_h \cdot I_w$. This strongly reduces complexity while providing equivalent uncertainty estimates to pixel-level sampling. 

Since the context patches obtained via a sliding window are strongly overlapping and thus highly correlated, we suggest limiting the number of required patches further such that $M << d^2$. Our argumentation is that two context patches shifted by a single pixel (in particular for a relatively large size of both the image and the context patch) will provide a virtually identical prediction signal for most pixels in the patch, thus introducing computations that do not provide a substantial benefit for inference. We thus introduce an additional stride parameter $s$ that defines the step size of the sliding window. This additionally reduces sampling complexity to $O(I_{h}/s \cdot I_{w}/s)$, i.e. a decrease by a factor of $s^2$ for arbitrary context size $d \times d$. However, since the number of context patches per pixel similarly decreases to $M \approx d^2/s^2$ - ignoring effects at the image borders, see below - there is a trade-off in terms of sampling rate and estimation quality. Note that extracting context patches with a sliding window violates the \IID assumption on Monte Carlo samples supporting our theoretical motivation in \autoref{subsec:uq-methods-patches}, since they are not sampled uniformly random. Nonetheless, we consider using the sliding window approach for its practical implementation value as a sufficiently close approximation.

\subsection{Distribution of context patch counts}
\label{app:patch_borders}

\begin{figure}[h]
    \centering
    \includegraphics[width=0.4\columnwidth]{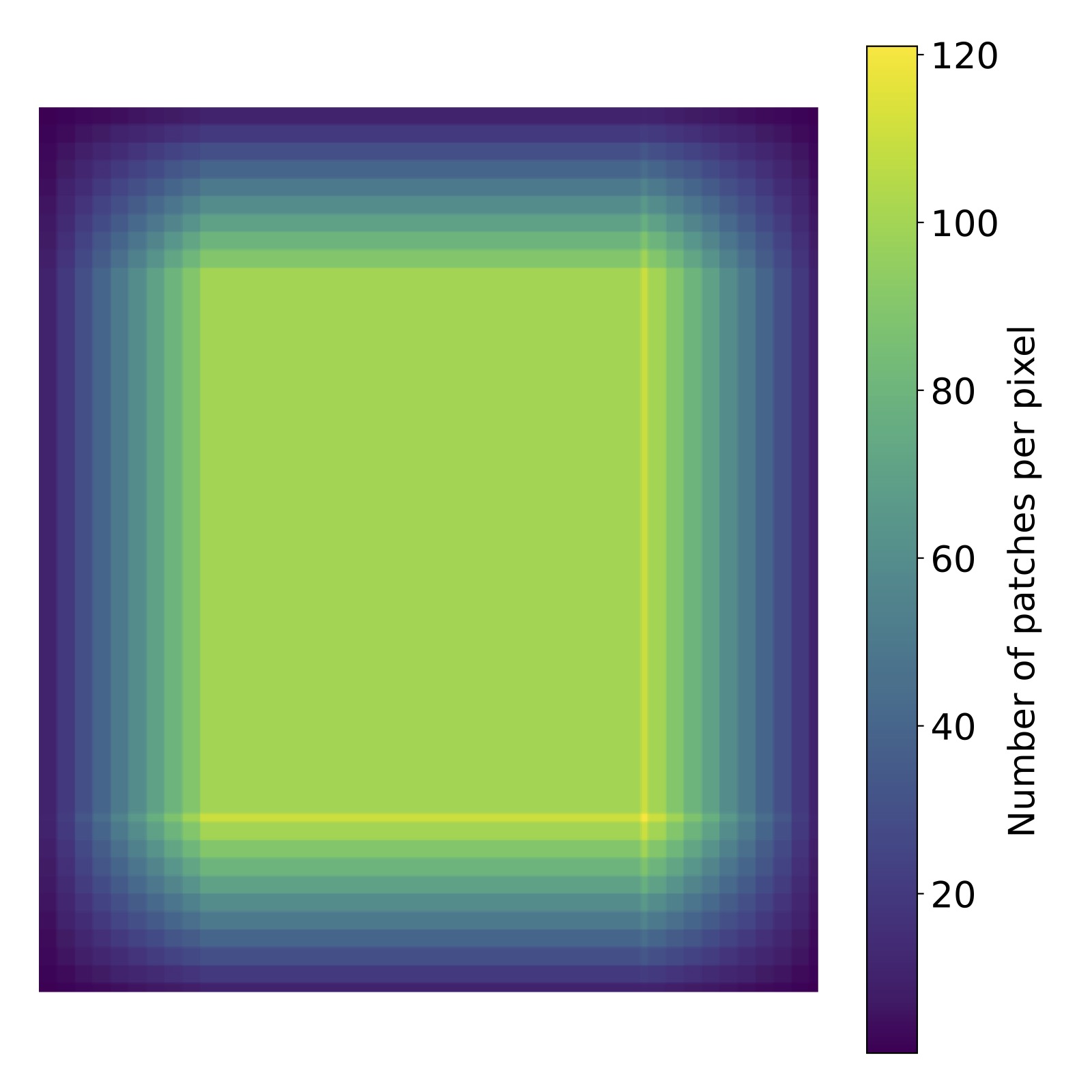}
    \caption{Empirical distribution of context patch counts $M$ per pixel in a gridded input image of size $495 \times 436$ for patches of size $d=100$ and stride $s=10$.}
    \label{fig:sample_number}
\end{figure}

In \autoref{app:patches_complexity} we state the expected sample count $M$ for a given pixel under a sliding window with stride $s$ to be $M \approx d^2/s^2$. In \autoref{fig:sample_number} we display the true recovered sample count per pixel in a gridded image of size $495 \times 436$ (i.e. our data size) using patches of size $d=100$ and stride $s=10$. We observe that the expected sample count $M = 100^2/10^2 = 100$ is satisfied for most of the centrally located pixels \footnote{Some pixels are in fact covered by even more than the expected number since we add an additional patch row and column to ensure that the right and bottom edges of the image are covered.}, but degrades as we move towards the image borders, resulting in an average sample count of $M=66.5$. This is due to the fact that we require a fixed patch size for model processing, and that image padding would feed the model with inaccurate snapshots of the current traffic state and thus not be beneficial in the context of UQ. In other words, context patches have to be fully located within the input image. 

Specifically, let us consider the closest distances of $\mathcal{P}$ located at $(i,j)$ to the image border $\mathcal{B}$ in vertical and horizontal direction respectively as $d_v(\mathcal{P}, \mathcal{B}) = \min\{ |1-i|,\,|I_h-i| \}+1,\,\,d_h(\mathcal{P}, \mathcal{B}) = \min\{ |1-j|,\,|I_w-j| \}+1$. Intuitively, the minimum number of `hops' from $\mathcal{P}$ to the closest border pixel is computed. The expected number of context patches sampled for $\mathcal{P}$ would then be $$ M=f\left(\frac{d_{v}(\mathcal{P}, \mathcal{B})}{s}\right) \cdot f\left(\frac{d_{h}(\mathcal{P}, \mathcal{B})}{s}\right),$$ where $f(x) = \lfloor x \rfloor$ if $x \ge 1$ and $f(x) = \lceil x \rceil$ if $x < 1$. Consider the previous example in \autoref{fig:theory-patch-ex} where we have $I_h = I_w = 9$ and placed $\mathcal{P}$ at $(i,j)=(5,5)$. In that case we obtain $d_v(\cdot)=d_h(\cdot)=5$, and by setting the stride to $s=1$ we would expect $M=5\cdot5=25$ context patches for $\mathcal{P}$, i.e. full coverage under every context patch sampled via the sliding window. For a pixel located at $(i,j)=(2,1)$ closer to the border we get $d_v(\cdot)=2,\,d_h(\cdot)=1$ and thus only $M=2\cdot1=2$. There is therefore a notable degradation in Monte Carlo sample counts $M$ as one moves towards the image border, negatively affecting the quality of obtained pixel-level uncertainty estimates. Of course, for the given example where $d$ is very large relative to $I_h$ and $I_w$ this degradation is particularly pronounced, while for large images and smaller patch sizes this issue is less critical.

\subsection{Effects of pixel centrality and stride parameter}
\label{app:patch_effects}

We additionally consider how the position of a pixel $\mathcal{P}$ within its context patch may affect inference, specifically in terms of predictive performance. Intuitively, we may expect that the prediction error as measured via MSE may be lower for pixels located centrally within patches, since the model is able to better take into account spatial traffic information from all directions. We compare the predictive performance of three aggregation methods for pixel-wise point estimates across context patches: \m{\#1:} standard averaging over patch-wise predictions; \m{\#2:} picking the prediction for the context patch in which $\mathcal{P}$ is most centrally located; \m{\#3:} a weighted average over patch-wise predictions, where weights are proportional to the pixel's centrality of positioning, i.e. the mean distance in both horizontal and vertical direction to the closest patch edges. We find that using both aggregation methods \m{\#2} and \m{\#3} slightly increase MSE scores, with \m{\#2} recording a stronger increase. That is, simple averaging performs best, suggesting that pixel centrality is not a major determining factor for predictive performance. However, future work could aim at optimizing the weights of patch-wise predictions with potential gains, as has been recently explored for test time augmentation \citep{shanmugam2021}.

We also consider varying the choice of stride parameter $s$. A lower stride results in higher context patch counts $M$, and should lead to improved performance. However, these patches will be highly correlated, so the trade-off between additional gain from higher sample counts and required inference time is unclear. We evaluate \m{Patches} using different stride values $s \in \{10, 20, 30, 40\}$. That is, choosing $s=30$ will shift the sliding window by 30 pixels before sampling the next patch. We keep patch sizes fixed at $d=100$. We observe in \autoref{tab:app-stride} that uncertainty estimation quality improves, but at the cost of much longer inference time. On the other hand, gains in predictive performance are marginal. We select $s=10$ for our main experiments, however a more thorough analysis on the effects of choosing $s$ may prove beneficial.

\begin{table*}[h]
    \centering
    \resizebox{0.7\textwidth}{!}
    {
        \begin{tabular}{r | r r r r r r}
        \toprule
        Stride $s$ &  Mean patch count $M$ &  Inference time (sec) $\downarrow$ & MSE $\downarrow$ & $\rho_{sp}$ $\uparrow$ & MPIW $\downarrow$ \\
        \midrule
        10 & 66.5 & 33.263 & \textbf{40.693} & \textbf{0.736} & \textbf{3.662} \\
        20 & 17.5 & 9.246 & 40.699 & 0.698 & 4.084 \\
        30 & 9.0 & 4.973 & 40.708 & 0.692 & 4.263 \\
        40 & 5.1 & \textbf{3.037} & 40.718 & 0.644 & 5.302 \\
        \bottomrule
        \end{tabular}
    }
    \caption{Impact of choosing stride $s$ on \m{Patches} in terms of inference time against predictive and uncertainty estimation improvements. A trade-off between inference time and estimation quality is apparent. Results are for Antwerp and from a previous series of experiments and should be considered indicative. Best scores marked in \textbf{bold}.}
    \label{tab:app-stride}
\end{table*}

\section{Unsupervised outlier detection: further details}
\label{app:outliers}

\begin{figure}[h]
	\includegraphics[width=\textwidth]{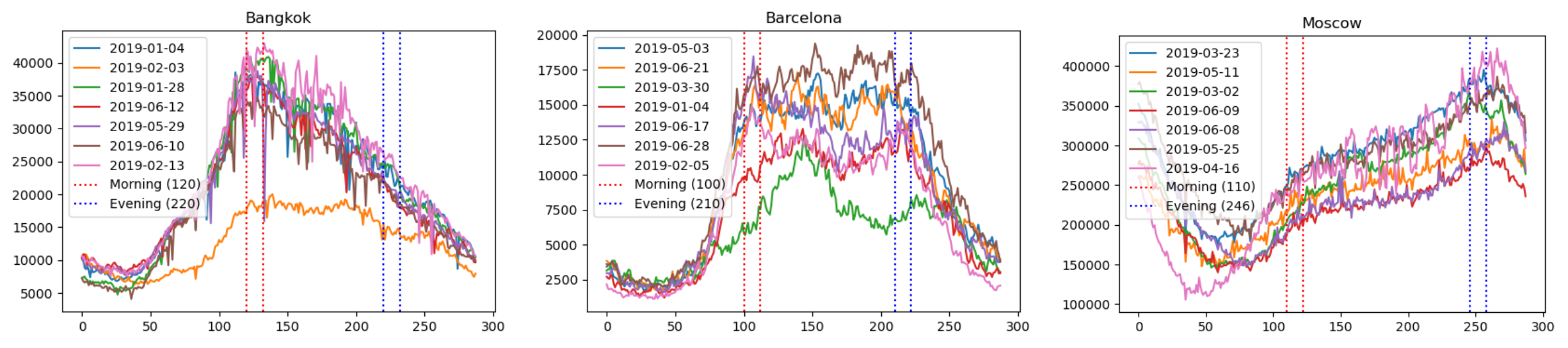}
	\centering
	\caption{Exemplary traffic activity time series for each city considered in our outlier detection framework. Traffic activity is measured as the sum over volume channels and displayed against time of day (288 samples at 5 min sample rate). Morning and evening rush hour intervals (12 steps starting from a fixed sample) are displayed and coincide with regions of high traffic activity. Selected fixed samples for each city (\m{from left to right}): 10:00 AM and 06:20 PM (indices 120 and 220); 08:20 AM and 05:30 PM (indices 100 and 210); 09:20 AM and 08:30 PM (indices 110 and 246).}
	\label{fig:app-ts-all}
\end{figure}

\begin{figure}[h]
	\includegraphics[width=0.8\textwidth]{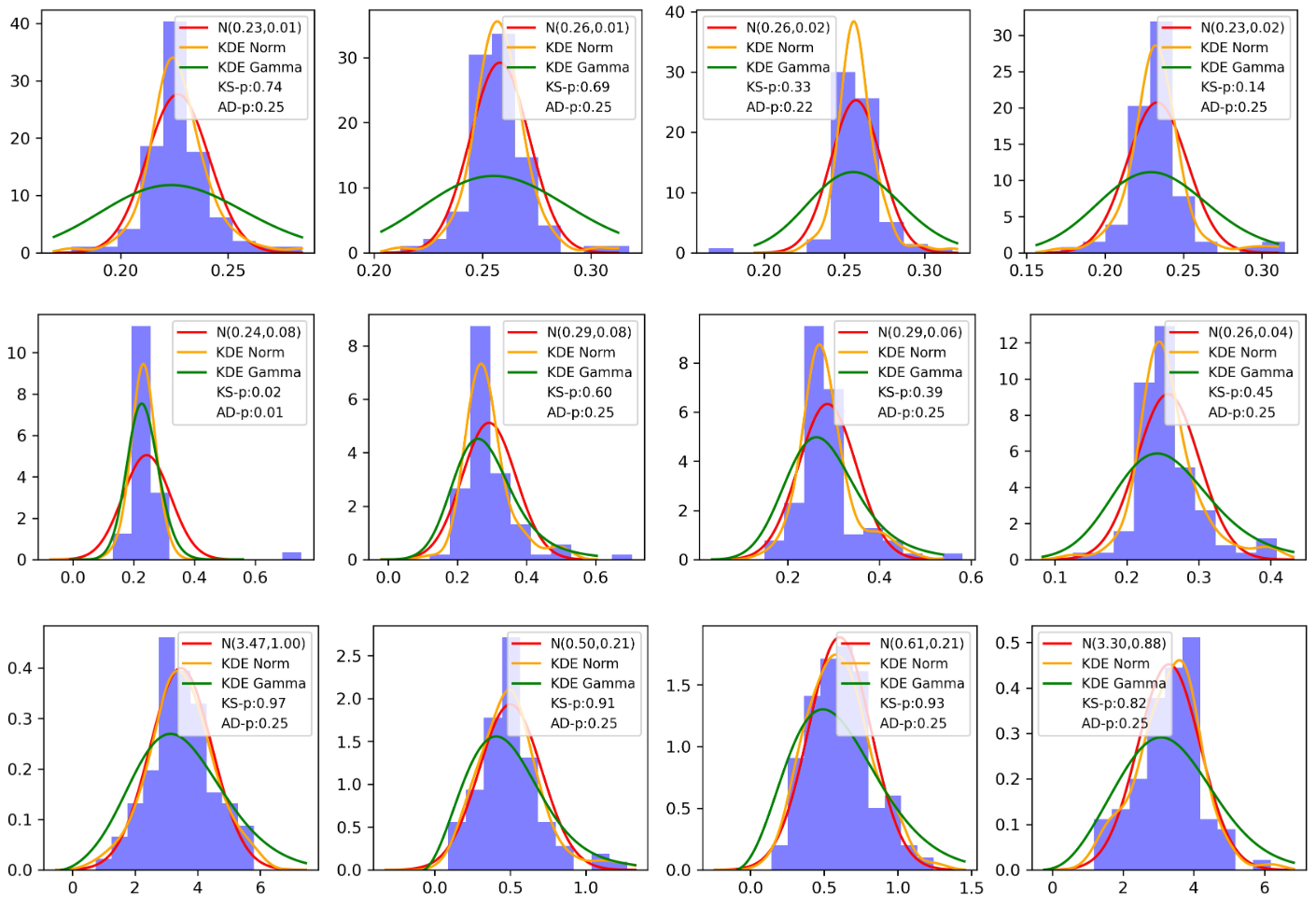}
	\centering
	\caption{Exemplary kernel density estimation (KDE) fits on channel-wise epistemic uncertainty estimates as specified in Algorithm \ref{alg:outlier_detection}. \m{From top to bottom (by row)}: Bangkok, pixel $(200, 200)$; Barcelona, pixel $(300, 300)$; Moscow, pixel $(300, 100)$. \m{KS-p} and \m{AD-p} are p-values for Kolmogorov-Smirnoff and Anderson-Darling tests on distribution fit for the Gaussian KDE fit (\m{KDE Norm}), which exhibits a good fit among options. Note that any \m{AD-p} $> 0.25$ are mapped to $0.25$.}
	\label{fig:app-kde-fits}
\end{figure}

\begin{figure}[h]
	\includegraphics[width=0.6\textwidth]{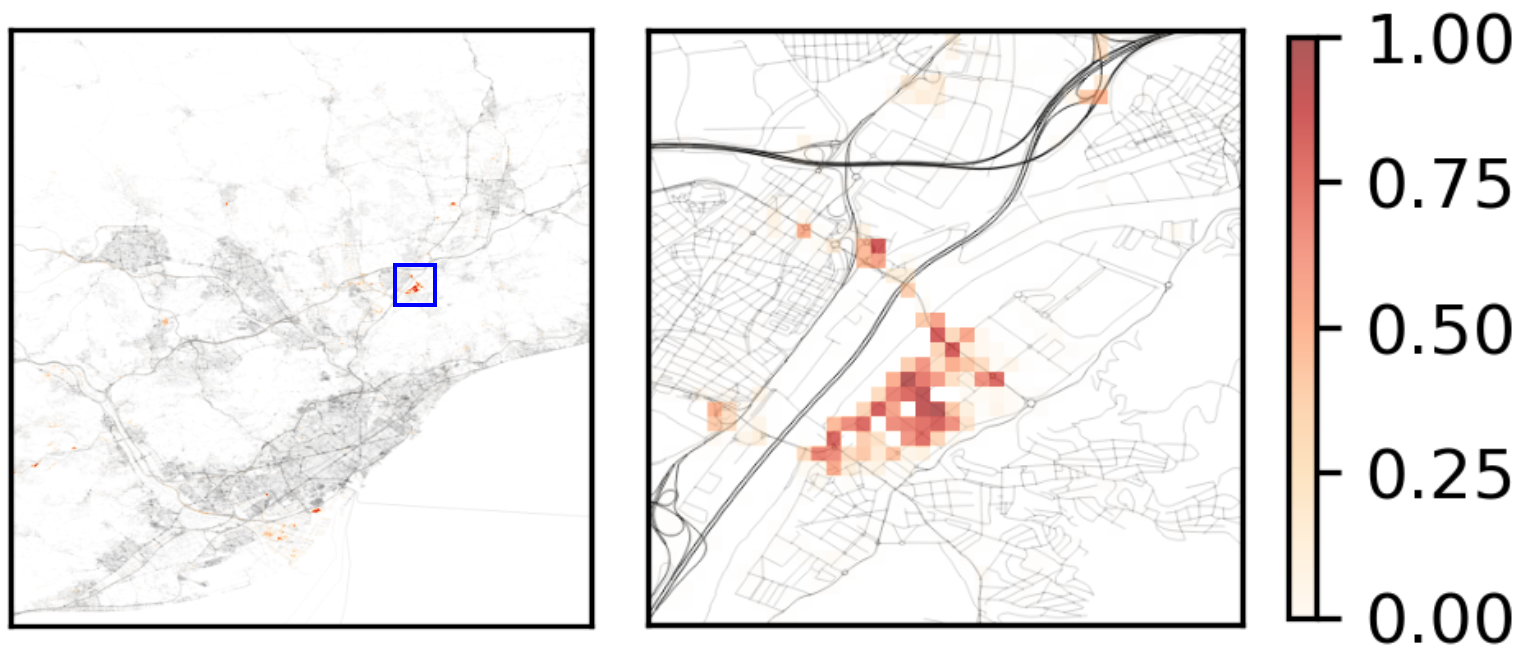}
	\centering
	\caption{Spatial distribution of pixel outliers for Barcelona, with focus on a specific city crop. This city crop corresponds to a business park on the outskirts of the city (street crossing \emph{Carrer de Sant Marti/Carrer Gorg}) and can be considered a spatial region with intrinsically high traffic activity.}
	\label{fig:app-out-barc-crop}
\end{figure}

\begin{figure}[h]
	\includegraphics[width=0.8\textwidth]{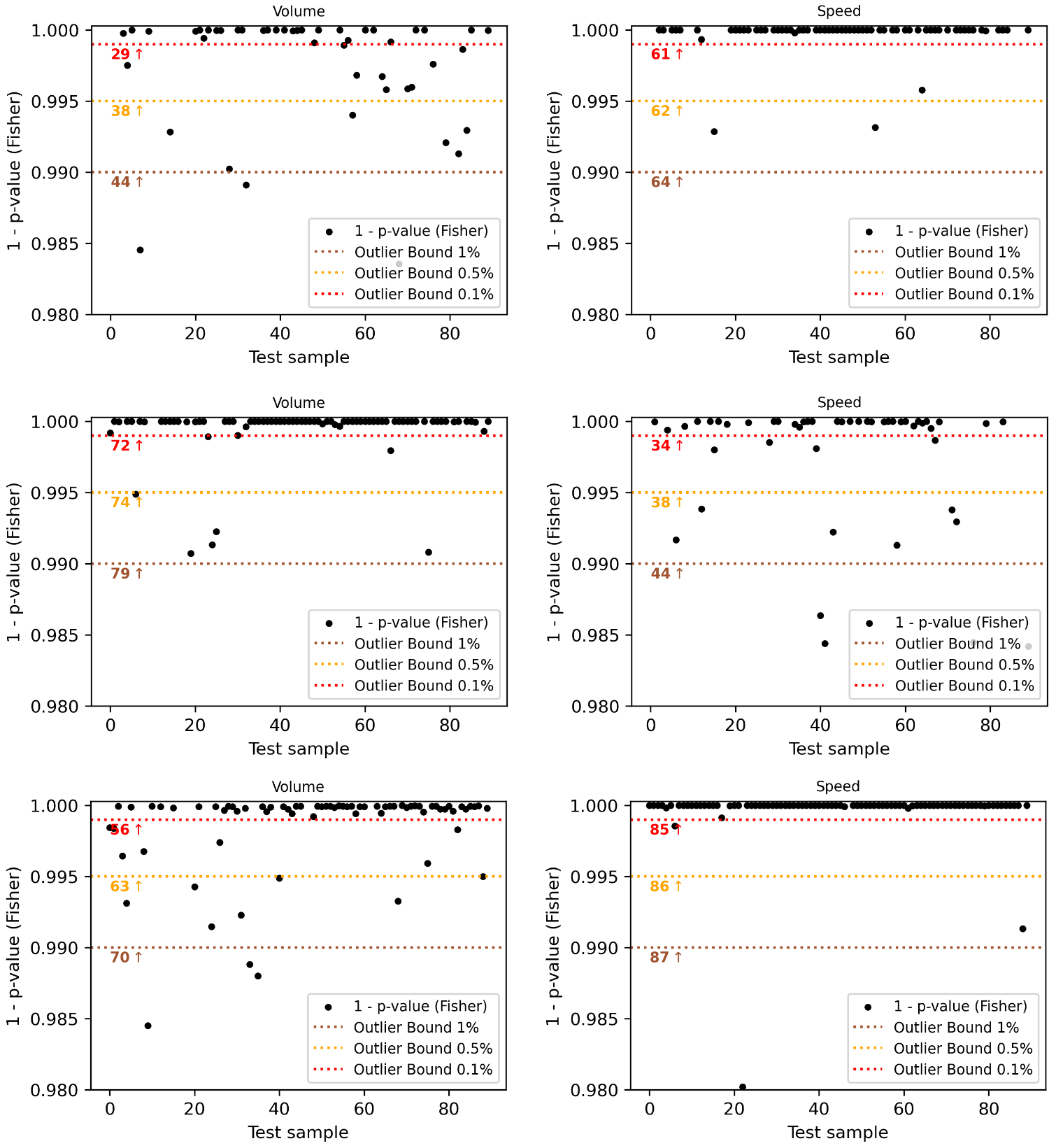}
	\centering
	\caption{Aggregated p-values (using Fisher's method) against different outlier bounds for volume (\m{left}) and speed (\m{right}) channel groups, for exemplary pixels with high positive outlier counts. \m{From top to bottom (by row)}: Bangkok, pixel $(461, 24)$; Barcelona, pixel $(210, 304)$; Moscow, pixel $(245, 405)$. We display (1 - p-value) and count outlier labels falling above a given threshold. We observe that most outliers are labelled positively regardless of choice of bound, since their p-values are highly significant.}
	\label{fig:app-out-pvalues}
\end{figure}

\begin{figure}[h]
	\includegraphics[width=0.8\textwidth]{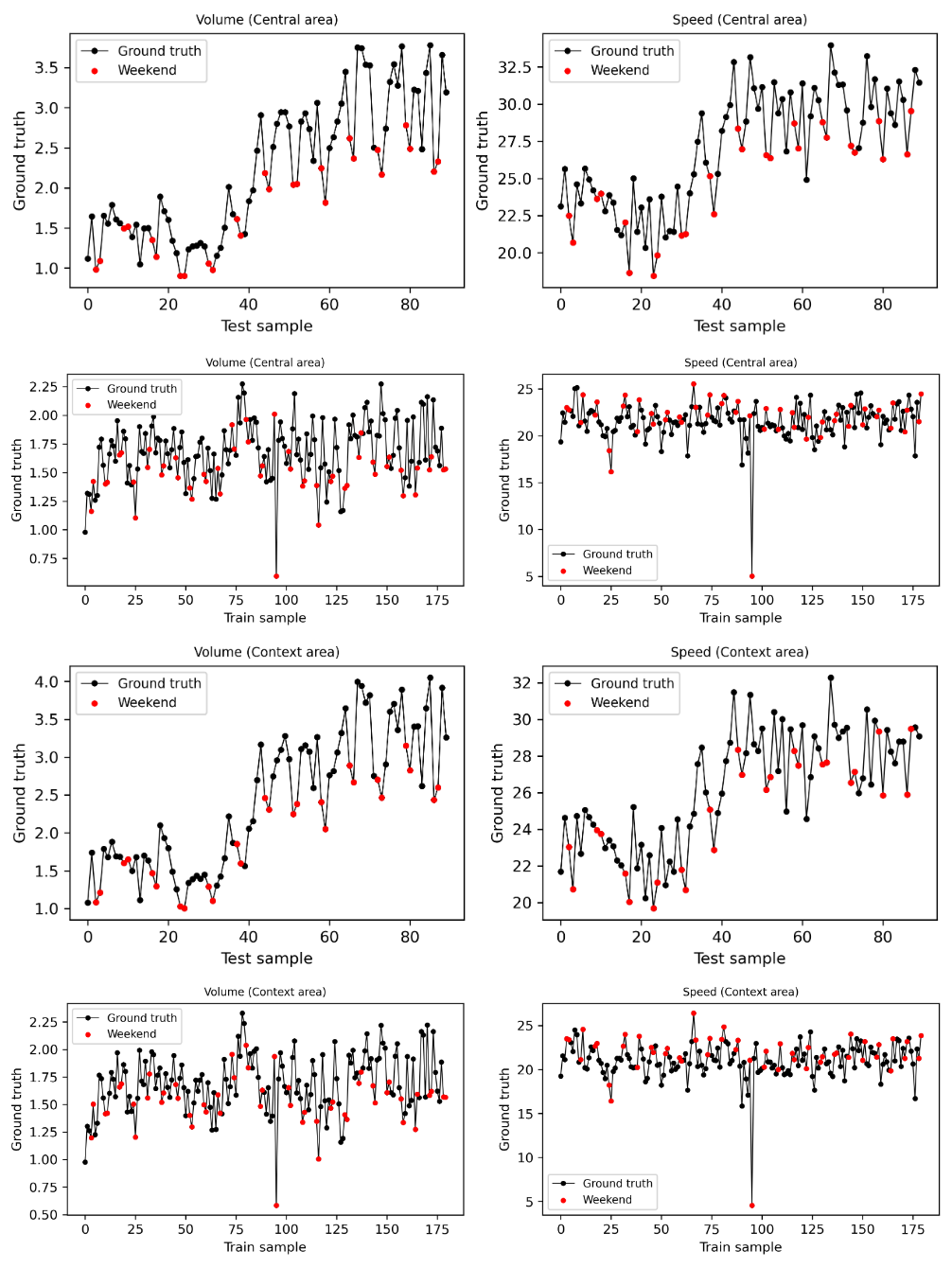}
	\centering
	\caption{Ground truth (GT) time series for selected city crop \m{\#1} in Moscow (Nagatinsky Zaton dist.), separated by volume (\m{left}) and speed channels (\m{right}). GT values are means across the channel group. \m{From top to bottom:} GT for the \emph{central} crop area on test samples, GT for the \emph{central} crop area on train samples, GT for the \emph{context} crop area on test samples, GT for the \emph{context} crop area on train samples.}
	\label{fig:app-out-crop1-moscow-ts}
\end{figure}

\begin{figure}[h]
	\includegraphics[width=0.8\textwidth]{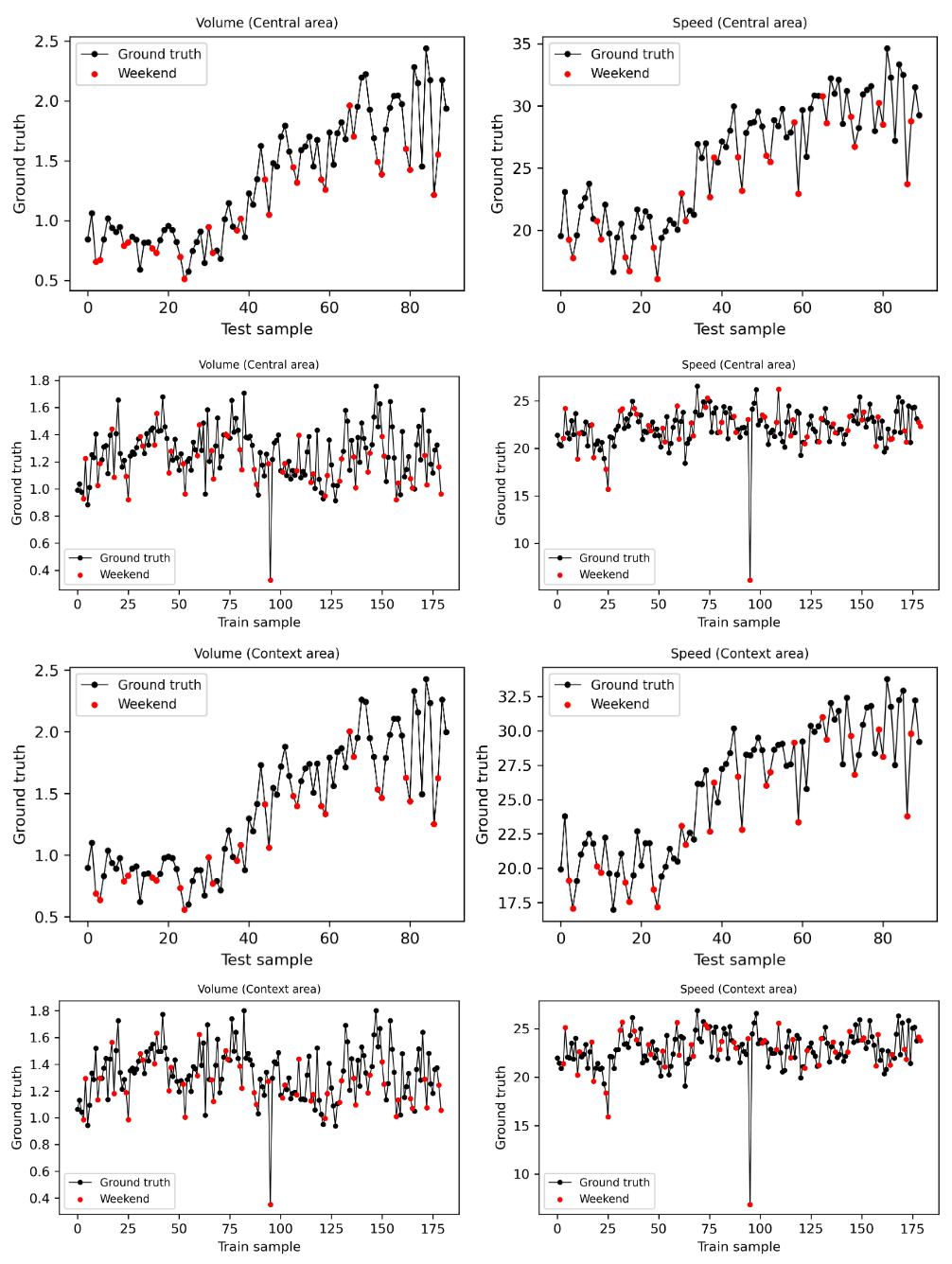}
	\centering
	\caption{Ground truth (GT) time series for selected city crop \m{\#2} in Moscow (Ramenki dist.), separated by volume (\m{left}) and speed channels (\m{right}). GT values are means across the channel group. \m{From top to bottom:} GT for the \emph{central} crop area on test samples, GT for the \emph{central} crop area on train samples, GT for the \emph{context} crop area on test samples, GT for the \emph{context} crop area on train samples.}
	\label{fig:app-out-crop2-moscow-ts}
\end{figure}

\begin{figure}[h]
	\includegraphics[width=0.8\textwidth]{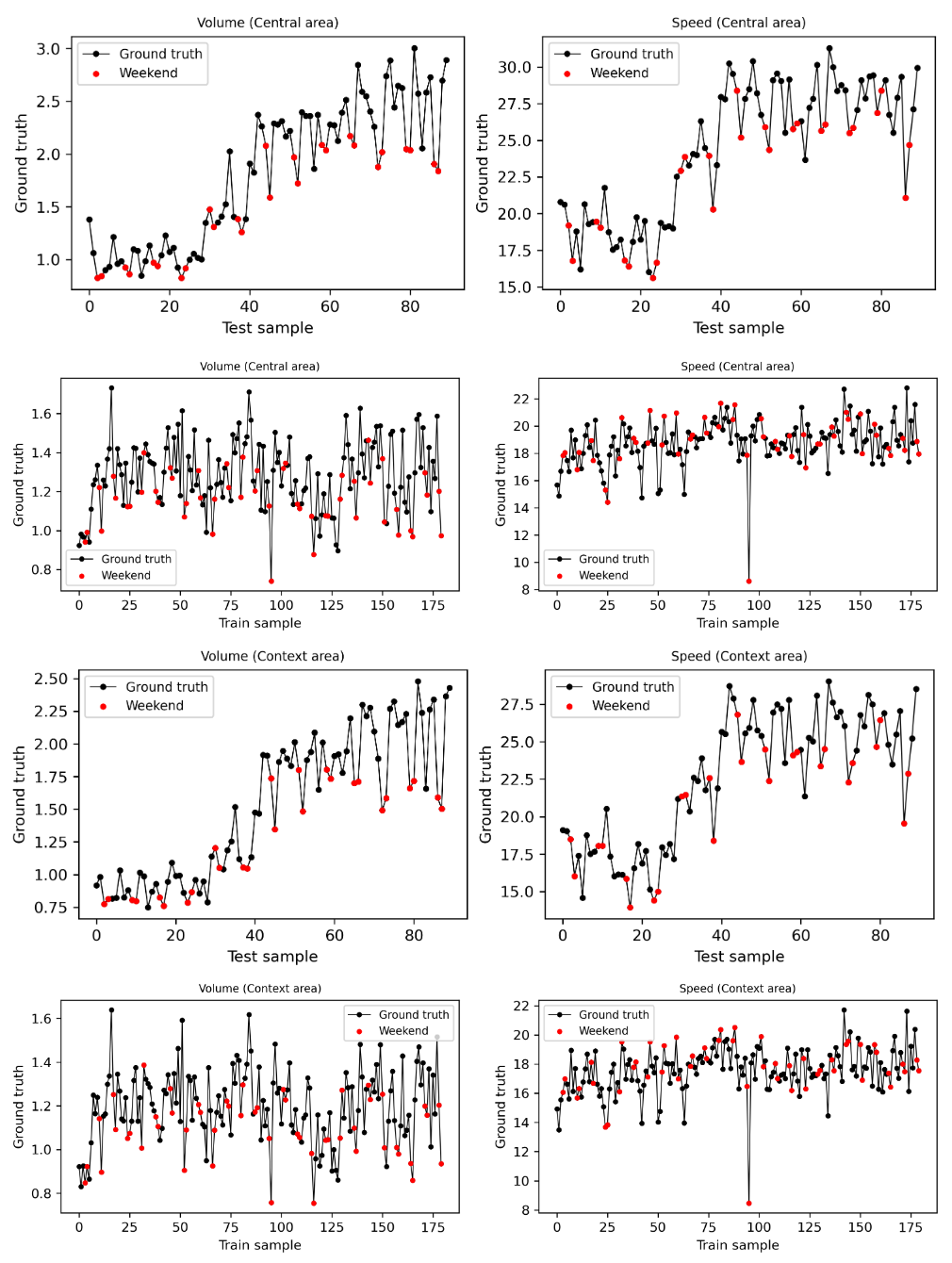}
	\centering
	\caption{Ground truth (GT) time series for selected city crop \m{\#1} in Moscow (Vladykino area), separated by volume (\m{left}) and speed channels (\m{right}). GT values are means across the channel group. \m{From top to bottom:} GT for the \emph{central} crop area on test samples, GT for the \emph{central} crop area on train samples, GT for the \emph{context} crop area on test samples, GT for the \emph{context} crop area on train samples.}
	\label{fig:app-out-crop3-moscow-ts}
\end{figure}

We visualise exemplary time series on each city's traffic activity in \autoref{fig:app-ts-all}, motivating our choice of fixing morning and evening rush hour indices which match times of generally high traffic activity. We further visualize in \autoref{fig:app-kde-fits} different kernel density estimation (KDE) fits on the epistemic uncertainty estimates as specified in Algorithm \ref{alg:outlier_detection}. We observe that a KDE with Gaussian kernel provides a good fit on the data, which is supported by statistical tests on the distribution fit. Regarding robustness of detected positive outlier labels to the choice of outlier bound \m{ob}, we observe in \autoref{fig:app-out-pvalues} for exemplary pixels with high outlier counts that obtained aggregated p-values are generally highly significant, and are well above any reasonable choice of outlier threshold. This motivates our claim that setting e.g. $\m{ob}=1\%$ vs. $\m{ob}=0.1\%$ will result in similar conclusions. 
We display a specific city crop for Barcelona in \autoref{fig:app-out-barc-crop}, which was referenced in \autoref{subsec:outliers-temporal}. Finally, we display the ground truth time series for the three city crops in Moscow, on which our interpretation in \autoref{subsec:outliers-spatial} is based on. See also that section on how to interpret these figures (\autoref{fig:app-out-crop1-moscow-ts}, \autoref{fig:app-out-crop2-moscow-ts}, \autoref{fig:app-out-crop3-moscow-ts}).

\FloatBarrier

\section{Uncertainty quantification: further results}
\label{app:results_table}

\begin{figure}[h]
	\includegraphics[width=0.95\textwidth]{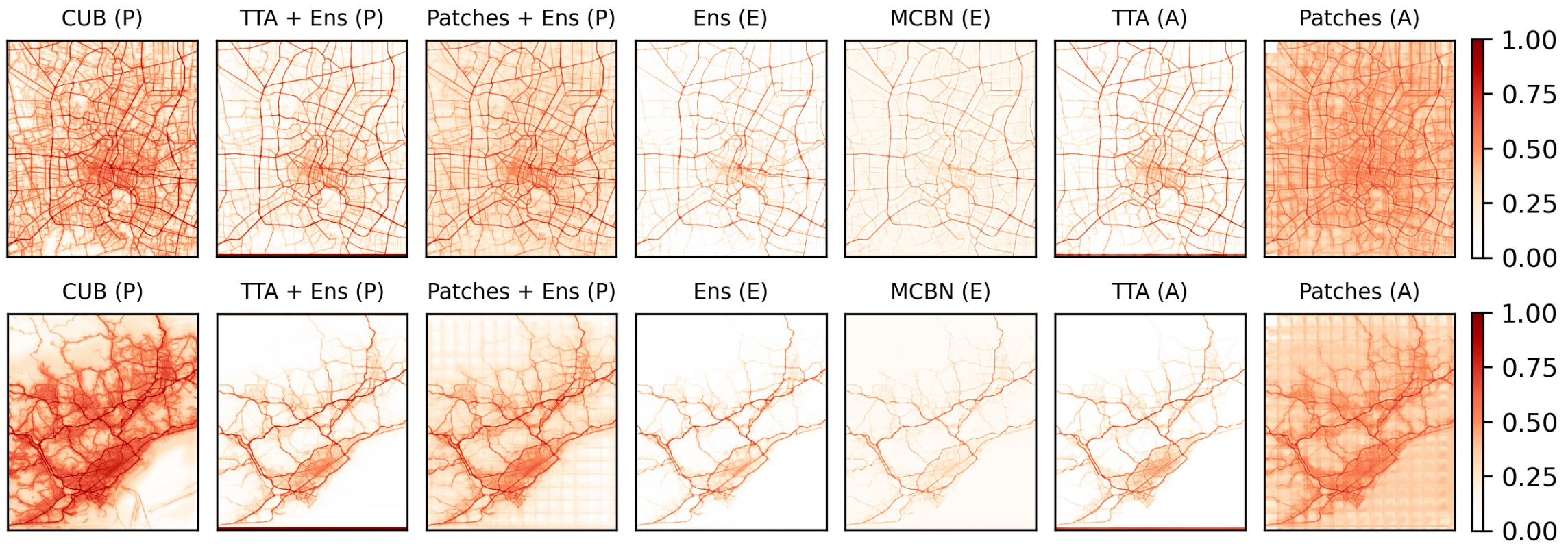}
	\centering
	\caption{Spatial distribution of uncertainty estimates for Bangkok (\m{top}) and Barcelona (\m{bottom}) across UQ methods. Values are logarithmised and normalized to $[0,1]$. Uncertainty values spatially recover the city's road network.}
	\label{fig:app-spatial-uq}
\end{figure}

\begin{figure}[h]
	\includegraphics[width=0.95\textwidth]{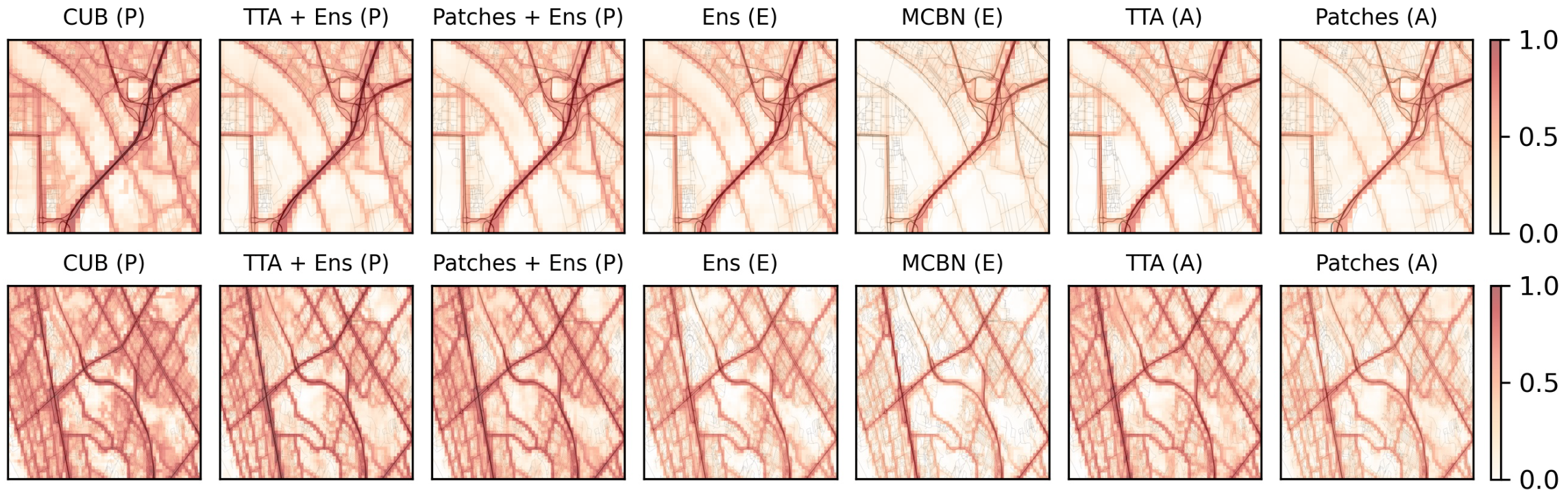}
	\centering
	\caption{Spatial distribution of uncertainty estimates for two city crops from Antwerp (\m{top}) and Moscow (\m{bottom}) across UQ methods. Values are logarithmised and normalized to $[0,1]$. Uncertainty values spatially recover the city's road network also at the more granular level.}
	\label{fig:app-spatial-uq-crop}
\end{figure}

We report additional results tables for the comparison of UQ methods. \autoref{tab:app_uq_unetpp} contains results for Bangkok and Barcelona using a U-Net++ model; \autoref{tab:app_uq_unet1} contains results for Antwerp and Moscow using a U-Net model; and \autoref{tab:app_uq_unet2} contains results for Bangkok and Barcelona using a U-Net model. We observe that across cities and methods the predictive uncertainty methods \m{TTA+Ens} and \m{Patches+Ens} typically perform the best. We also display the spatial distribution of uncertainty across UQ methods on Bangkok and Barcelona in \autoref{fig:app-spatial-uq}, and on sample city crops from Antwerp and Moscow in \autoref{fig:app-spatial-uq-crop}. We observe that spatially coherent uncertainty estimates are recovered.

\begin{table*}[t]
\centering
\resizebox{\textwidth}{!}{

\begin{tabular}{l|rrrrr|rrrrr}

\toprule
{\textbf{City}} & \multicolumn{5}{c}{Bangkok (temporal transfer)} & \multicolumn{5}{c}{Barcelona (temporal transfer)} \\

\midrule

\textbf{UQ method} & MSE $\downarrow$ & Uncertainty & MPIW $\downarrow$ & ENCE $\downarrow$ & $\rho_{sp}$ $\uparrow$
& MSE $\downarrow$ & Uncertainty & MPIW $\downarrow$ & ENCE $\downarrow$ & $\rho_{sp}$ $\uparrow$ \\

\midrule

\m{Ens (E)} & \textbf{55.19} & 0.142 $\pm$ 0.092 & 1.450 & 0.645 & 0.630 & 33.52 & 0.108 $\pm$ 0.082 & 0.752 & 0.552 & 0.461 \\
$\hookrightarrow$ \m{zero-mask}  & \textbf{224.43} & 0.440  $\pm$ 0.269 & 5.319 & 1.769 & 0.755 & 201.51 & 0.450  $\pm$  0.360 & 3.834 & 1.354 & 0.816 \\

\m{MCBN (E)} & 56.00 & 0.134 $\pm$ 0.097 & 2.520 & 1.396 & 0.655 & 33.28 & 0.100 $\pm$ 0.084 & 1.381 & 0.751 & 0.792 \\
$\hookrightarrow$ \m{zero-mask}  & 227.61 & 0.353 $\pm$ 0.257 & 8.796 & 3.287 & 0.678  & 199.99 & 0.301 $\pm$ 0.298 & 6.639 & 2.738 & 0.714 \\

\midrule

\m{TTA (A)} & 56.46 & 0.148 $\pm$ 0.102 & 1.389 & 0.940 & 0.368 & 33.40 & 0.089 $\pm$ 0.067 & 0.647 & 0.912 & 0.236 \\
$\hookrightarrow$ \m{zero-mask}   & 229.55 & 0.422 $\pm$ 0.304 & 5.029 & 1.986 & 0.625  & 200.82 & 0.305 $\pm$ 0.307 & 3.330 & 1.849 & 0.684 \\

\m{Patches (A)} & 55.66 & 0.049 $\pm$  0.040 & 2.905 & 6.504 & 0.687 & 33.40 & 0.037 $\pm$ 0.036 & 1.581 & 5.682 & 0.511 \\
$\hookrightarrow$ \m{zero-mask}  & 226.31 & 0.147 $\pm$ 0.113 & 10.854 & 10.310 & 0.746  & 200.78 & 0.164 $\pm$ 0.165 & 8.525 & 6.711 & 0.799 \\

\midrule

\m{TTA+Ens (P)} & 55.41 & 0.671 $\pm$  0.260 & 1.209 & 0.464 & 0.797 & \textbf{33.07} & 0.533 $\pm$ 0.179 & 0.579 & 0.525 & 0.772 \\
$\hookrightarrow$ \m{zero-mask}  & 225.32 & 1.367 $\pm$ 0.706 & 4.392 & \textbf{0.595} & 0.834  & \textbf{198.83} & 0.968 $\pm$ 0.748 & 2.904 & 0.504 & 0.888 \\

\m{Patches+Ens (P)} & 57.10 & 0.212 $\pm$ 0.139 & \textbf{0.910} & \textbf{0.259} & \textbf{0.916} & 33.41 & 0.141 $\pm$ 0.118 & \textbf{0.549} & \textbf{0.277} & \textbf{0.887} \\
$\hookrightarrow$ \m{zero-mask}  & 232.20 & 0.703 $\pm$ 0.418 & \textbf{3.275} & 0.641 & \textbf{0.926}  & 200.91 & 0.658 $\pm$ 0.536 & \textbf{2.812} & \textbf{0.447} & \textbf{0.946} \\

\m{CUB (P)} & 56.31 & 0.325 $\pm$ 0.0 & 3.142 & 0.914 & 0.012 & 33.45 & 0.249 $\pm$ 0.0 & 1.910 & 6.235 & 0.029 \\
$\hookrightarrow$ \m{zero-mask}  & 228.95 & 1.136 $\pm$ 0.0 & 12.085 & 1.045 & 0.014  & 201.11 & 1.290  $\pm$ 0.0 & 10.780 & 1.010 & 0.026 \\

\bottomrule

\end{tabular}
}
\vspace{1mm}
    \caption{Comparison of UQ methods for \textbf{Bangkok} and \textbf{Barcelona} in terms of traffic speed prediction. Predictions are obtained using a \textbf{U-Net++} model. Methods are grouped by modelled uncertainty (\m{E}: epistemic, \m{A}: aleatoric, \m{P}: predictive). Scores are calculated both across all cells and only those cells with traffic activity, i.e. non-zero volume (\m{zero-mask}). Best scores per city, metric and masking across all methods are marked in \textbf{bold}. Arrows indicate favoured score direction (higher $\uparrow$, lower $\downarrow$).}
    \label{tab:app_uq_unetpp}
\end{table*}

\begin{table*}[t]
\centering
\resizebox{\textwidth}{!}{

\begin{tabular}{l|rrrrr|rrrrr}

\toprule
{\textbf{City}} & \multicolumn{5}{c}{Antwerp (spatio-temporal transfer)} & \multicolumn{5}{c}{Moscow (temporal transfer)} \\

\midrule

\textbf{UQ method} & MSE $\downarrow$ & Uncertainty & MPIW $\downarrow$ & ENCE $\downarrow$ & $\rho_{sp}$ $\uparrow$
& MSE $\downarrow$ & Uncertainty & MPIW $\downarrow$ & ENCE $\downarrow$ & $\rho_{sp}$ $\uparrow$ \\

\midrule

\m{Ens (E)} & 80.83 & 0.379 $\pm$ 0.162 & 1.427 & 0.958 & 0.262 & 198.10 & 0.655 $\pm$ 0.247 & 16.930 & 2.661 & 0.195 \\
$\hookrightarrow$ \m{zero-mask} & 298.70 & 0.675 $\pm$  0.440 & 4.570 & 1.428 & 0.454 & 324.40 & 0.879 $\pm$  0.360 & 27.510 & 4.127 & 0.278 \\

\m{MCBN (E)} & 80.53 & 0.272 $\pm$ 0.318 & 3.098 & 1.131 & 0.159 & 204.40 & 0.425 $\pm$ 0.215 & 18.410 & 4.322 & 0.259 \\
$\hookrightarrow$ \m{zero-mask} & 297.60 & 0.646 $\pm$ 0.587 & 11.210 & 1.691 & 0.387 & 334.70 & 0.661 $\pm$ 0.323 & 30.010 & 7.072 & 0.382 \\

\midrule

\m{TTA (A)} & 79.840 & 0.294 $\pm$ 0.244 & 2.696 & 1.312 & 0.258 & 198.20 & 0.996 $\pm$ 0.403 & 16.980 & 1.910 & 0.234 \\
$\hookrightarrow$ \m{zero-mask} & 295.00 & 0.662 $\pm$ 0.568 & 9.721 & 2.165 & 0.444 & 324.60 & 1.406 $\pm$ 0.564 & 27.650 & 2.625 & 0.344 \\

\m{Patches (A)} & \textbf{77.18} & 0.041 $\pm$ 0.042 & 12.78 & 40.42 & 0.394 & 196.2 & 0.13  $\pm$ 0.083 & 83.26 & 40.98 & 0.458 \\
$\hookrightarrow$ \m{zero-mask} & \textbf{284.8} & 0.101 $\pm$ 0.105 & 36.02 & 39.74 & 0.615 & 321.2 & 0.199 $\pm$ 0.126 & 132.0 & 45.18 & 0.528 \\

\midrule

\m{TTA+Ens (P)} & 81.57 & 0.790  $\pm$  0.390 & 1.196 & 0.400 & 0.793 & 206.50 & 1.789 $\pm$ 0.574 & 15.720 & \textbf{0.921} & 0.633  \\
$\hookrightarrow$ \m{zero-mask} & 301.40 & 1.350  $\pm$ 0.913 & 3.546 & \textbf{0.591} & 0.842 & 338.20 & 2.423 $\pm$ 0.814 & 25.470 & \textbf{1.283} & 0.660 \\

\m{Patches+Ens (P)} & 82.990 & 0.332 $\pm$ 0.215 & \textbf{1.069} & \textbf{0.336} & \textbf{0.856} & 218.9 & 1.105 $\pm$ 0.386 & 15.620 & 1.150 & \textbf{0.714} \\
$\hookrightarrow$ \m{zero-mask} & 306.70 & 0.788 $\pm$ 0.585 & \textbf{3.189} & 0.669 & \textbf{0.891} & 358.50 & 1.663 $\pm$ 0.582 & 25.350 & 1.811 & \textbf{0.701} \\

\m{CUB (P)} & 79.470 & 0.572 $\pm$ 0.0 & 3.941 & 1.089 & 0.770 & \textbf{195.80} & 1.696 $\pm$ 0.0 & \textbf{14.070} & 1.223 & 0.455 \\
$\hookrightarrow$ \m{zero-mask}  & 293.60 & 1.936 $\pm$ 0.0 & 14.200 & 1.107 & 0.432 & \textbf{320.60} & 2.722 $\pm$ 0.0 & \textbf{22.900} & 1.398 & 0.212 \\

\bottomrule

\end{tabular}
}
\vspace{1mm}
    \caption{Comparison of UQ methods for \textbf{Antwerp} and \textbf{Moscow} in terms of traffic speed prediction. Predictions are obtained using a \textbf{U-Net} model. Methods are grouped by modelled uncertainty (\m{E}: epistemic, \m{A}: aleatoric, \m{P}: predictive). Scores are calculated both across all cells and only those cells with traffic activity, i.e. non-zero volume (\m{zero-mask}). Best scores per city, metric and masking across all methods are marked in \textbf{bold}. Arrows indicate favoured score direction (higher $\uparrow$, lower $\downarrow$).}
    \label{tab:app_uq_unet1}
\end{table*}

\begin{table*}[t]
\centering
\resizebox{\textwidth}{!}{

\begin{tabular}{l|rrrrr|rrrrr}

\toprule
{\textbf{City}} & \multicolumn{5}{c}{Bangkok (temporal transfer)} & \multicolumn{5}{c}{Barcelona (temporal transfer)} \\

\midrule

\textbf{UQ method} & MSE $\downarrow$ & Uncertainty & MPIW $\downarrow$ & ENCE $\downarrow$ & $\rho_{sp}$ $\uparrow$
& MSE $\downarrow$ & Uncertainty & MPIW $\downarrow$ & ENCE $\downarrow$ & $\rho_{sp}$ $\uparrow$ \\

\midrule

\m{Ens (E)} & \textbf{55.64} & 0.329 $\pm$ 0.080 & 1.141 & 1.003 & 0.131 & \textbf{32.95} & 0.313 $\pm$ 0.073 & 0.601 & 0.964 & 0.129 \\
$\hookrightarrow$ \m{zero-mask} & \textbf{226.2} & 0.553 $\pm$ 0.225 & 4.218 & 1.619 & 0.272 & \textbf{198.10} & 0.610  $\pm$ 0.316 & 3.189 & 1.222 & 0.368 \\

\m{MCBN (E)} & 56.22 & 0.183 $\pm$ 0.182 & 2.887 & 1.165 & 0.092 & 33.18 & 0.327 $\pm$ 0.531 & 1.582 & 1.034 & 0.031 \\
$\hookrightarrow$ \m{zero-mask}  & 228.60 & 0.507 $\pm$ 0.366 & 11.620 & 1.818 & 0.276 & 199.30 & 0.643 $\pm$ 0.626 & 9.354 & 1.386 & 0.252 \\

\midrule

\m{TTA (A)} & 56.30 & 0.262 $\pm$ 0.177 & 2.587 & 1.269 & 0.226 & 33.18 & 0.131 $\pm$ 0.091 & 1.097 & 1.188 & 0.612 \\
$\hookrightarrow$ \m{zero-mask}  & 228.9 & 0.530  $\pm$ 0.389 & 10.400 & 2.105 & 0.355 & 199.50 & 0.327 $\pm$ 0.335 & 6.480 & 2.126 & 0.409 \\

\m{Patches (A)} & 56.28 & 0.040  $\pm$ 0.036 & 11.800 & 33.730 & 0.381 & 33.46 & 0.040  $\pm$  0.030 & 5.995 & 28.010 & 0.170 \\
$\hookrightarrow$ \m{zero-mask}  & 228.50 & 0.111 $\pm$ 0.099 & 35.430 & 29.320 & 0.563 & 200.80 & 0.097 $\pm$ 0.108 & 23.840 & 24.200 & 0.558 \\

\midrule

\m{TTA+Ens (P)} & 56.43 & 0.746 $\pm$ 0.234 & 0.957 & 0.404 & 0.752 & 33.20 & 0.615 $\pm$ 0.152 & 0.501 & 0.404 & 0.658 \\
$\hookrightarrow$ \m{zero-mask}  & 229.40 & 1.100 $\pm$ 0.545 & 3.117 & \textbf{0.618} & 0.802 & 199.60 & 0.800 $\pm$ 0.581 & \textbf{2.012} & \textbf{0.513} & 0.852 \\

\m{Patches+Ens (P)} & 57.59 & 0.270  $\pm$ 0.125 & \textbf{0.901} & \textbf{0.327} & \textbf{0.814} & 33.71 & 0.201 $\pm$ 0.100 & \textbf{0.493} & \textbf{0.311} & \textbf{0.802} \\
$\hookrightarrow$ \m{zero-mask}  & 234.20 & 0.670  $\pm$  0.360 & \textbf{2.946} & 0.691 & \textbf{0.848} & 202.70 & 0.616 $\pm$ 0.448 & 2.030 & 0.516 & \textbf{0.889} \\

\m{CUB (P)} & 55.83 & 0.306$\pm$ 0.0 & 3.473 & 1.080 & 0.855 & 33.13 & 0.230  $\pm$ 0.0 & 1.750 & 1.109 & 0.916 \\
$\hookrightarrow$ \m{zero-mask}  & 227.00 & 1.162 $\pm$ 0.0 & 13.960 & 1.152 & 0.571 & 199.20 & 1.291 $\pm$ 0.0 & 10.330 & 1.137 & 0.618 \\

\bottomrule

\end{tabular}
}
\vspace{1mm}
    \caption{Comparison of UQ methods for \textbf{Bangkok} and \textbf{Barcelona} in terms of traffic speed prediction. Predictions are obtained using a \textbf{U-Net} model. Methods are grouped by modelled uncertainty (\m{E}: epistemic, \m{A}: aleatoric, \m{P}: predictive). Scores are calculated both across all cells and only those cells with traffic activity, i.e. non-zero volume (\m{zero-mask}). Best scores per city, metric and masking across all methods are marked in \textbf{bold}. Arrows indicate favoured score direction (higher $\uparrow$, lower $\downarrow$).}
    \label{tab:app_uq_unet2}
\end{table*}

\end{document}